%% file: main.tex
\pdfoutput=1
\documentclass[10pt,twocolumn,letterpaper]{article}

\usepackage[pagenumbers]{cvpr}      

\input{preamble}

\definecolor{cvprblue}{rgb}{0.21,0.49,0.74}
\usepackage[pagebackref,breaklinks,colorlinks,allcolors=cvprblue]{hyperref}
\usepackage{comment}
\usepackage{xcolor}         
\usepackage{tabularx}
\usepackage{graphicx}
\usepackage{xspace}
\usepackage{nicematrix}

\usepackage{multirow}
\usepackage{colortbl}
\usepackage{adjustbox}
\usepackage{siunitx}
\usepackage{rotating}

\RequirePackage[labelfont=bf,font=small,tableposition=bottom]{caption}
\RequirePackage[skip=3pt]{subcaption}

\newcommand{\dataset}{ICQ-Highlight\xspace}
\newcommand{\benchmark}{ICQ\xspace}

\newcommand{\modelnumber}{12\xspace}

\newcommand{\scribble}{\texttt{scribble}\xspace}
\newcommand{\realistic}{\texttt{realistic}\xspace}
\newcommand{\cinematic}{\texttt{cinematic}\xspace}
\newcommand{\cartoon}{\texttt{cartoon}\xspace}

\newcommand{\captioning}{MQ-Cap\xspace}
\newcommand{\summarization}{MQ-Sum\xspace}
\newcommand{\visquery}{VQ-Enc\xspace}
\newcommand{\suit}{SUIT}
\newcommand{\sumsuit}{MQ-Sum(+SUIT)\xspace}

\title{Localizing Events in Videos with Multimodal Queries}

\author{%
  Gengyuan Zhang $^{1,4}\thanks{Equal contribution}$
  \qquad\  Mang Ling Ada Fok $^{2}\footnotemark[1]$
  \qquad\  Jialu Ma $^{1}$
  \qquad\  Yan Xia $^{2,4}$ \\
 Daniel Cremers $^{2,4}$ 
 \qquad\ Philip Torr $^{3}$
  \qquad\ Volker Tresp $^{1,4}$
  \qquad\  Jindong Gu $^{3}$  \\ 
  $^1$ LMU Munich \qquad\
  $^2$ TU Munich \qquad\
  $^3$ University of Oxford \vspace{0.05cm} \\
  $^4$ Munich Center for Machine Learning (MCML) \\
  \texttt{zhang@dbs.ifi.lmu.de}
  \qquad\ \texttt{ada.fok@tum.de} 
  }

\begin{document}
\maketitle
\input{sec/0_abstract}    
\input{sec/1_intro}

\input{sec/2_related}
\input{sec/3_benchmark}

\input{sec/4_experiment}

\input{sec/5_conclusion}

{
    \small
    \bibliographystyle{ieeenat_fullname}
    \bibliography{main}
}

\input{sec/X_suppl}

\end{document}

%% file: preamble.tex
\def\eg{\emph{e.g}\onedot} 
\def\ie{\emph{i.e}\onedot} 
 
\def\etc{\emph{etc}\onedot} \def\vs{\emph{vs}\onedot}

\definecolor{comlecolor}{HTML}{86B5A1}
\definecolor{correcolor}{HTML}{B85A58}

\newcommand{\qm}{$q_m$\xspace}
\newcommand{\vref}{$v_{ref}$\xspace}
\newcommand{\tmod}{$t_{ref}$\xspace}
\newcommand{\rel}[1]{\tiny{(#1)}}

%% file: sec/0_abstract.tex
\begin{abstract}
Localizing events in videos based on semantic queries is a pivotal task in video understanding, with the growing significance of user-oriented applications like video search.
Yet, current research predominantly relies on natural language queries (NLQs), overlooking the potential of using multimodal queries (MQs) that integrate images to more flexibly represent semantic queries— especially when it is difficult to express non-verbal or unfamiliar concepts in words.
To bridge this gap, we introduce \benchmark, a new benchmark designed for localizing events in videos with MQs, alongside an evaluation dataset \dataset. 
To accommodate and evaluate existing video localization models for this new task, we propose 3 Multimodal Query Adaptation methods and a novel Surrogate Fine-tuning on pseudo-MQs strategy.
\benchmark systematically benchmarks \modelnumber state-of-the-art backbone models, spanning from specialized video localization models to Video LLMs, across diverse application domains.
Our experiments highlight the high potential of MQs in real-world applications.
We believe this benchmark is a first step toward advancing MQs in video event localization\footnote[1]{Our project is available at \url{https://icq-benchmark.github.io/}}.
\end{abstract}

%% file: sec/1_intro.tex
\section{Introduction}
\label{sec:intro}
\label{sec:intro}

Localizing semantic events in videos has been a long-standing task in the field of video understanding~\cite{yu2024self,yang2023vid2seq, zhang2023multi, krishna2017dense, pan2023scanning,ren2023timechat,bhattacharyyalook}.
User-centric applications like streaming media and short video platforms highlight the need to parse video segments for video search and video highlight/moment recommendations given user queries.

\begin{figure}[t!]
\begin{subfigure}{0.48\textwidth}
        \centering
        \includegraphics[width=.98\linewidth]{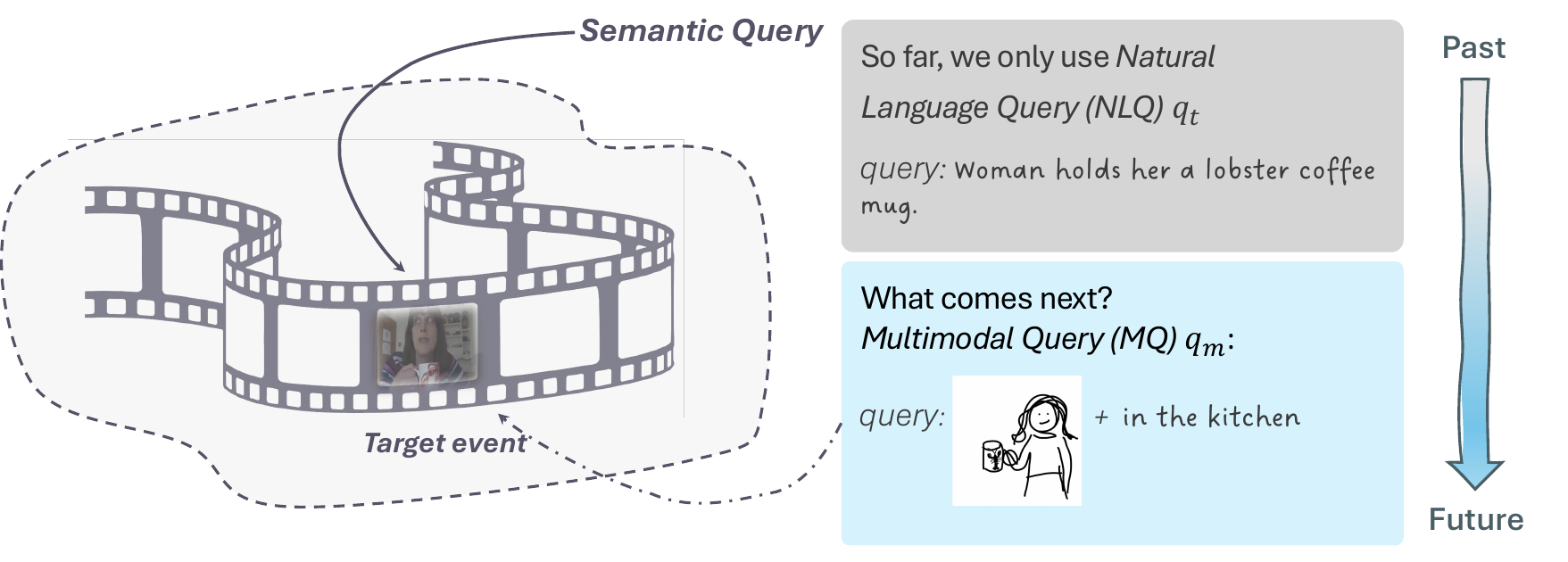}
        \caption{}
        \label{subfig:a}
    \end{subfigure}
    \hfill
    \begin{subfigure}{0.48\textwidth}
        \centering
        \includegraphics[width=.98\linewidth]{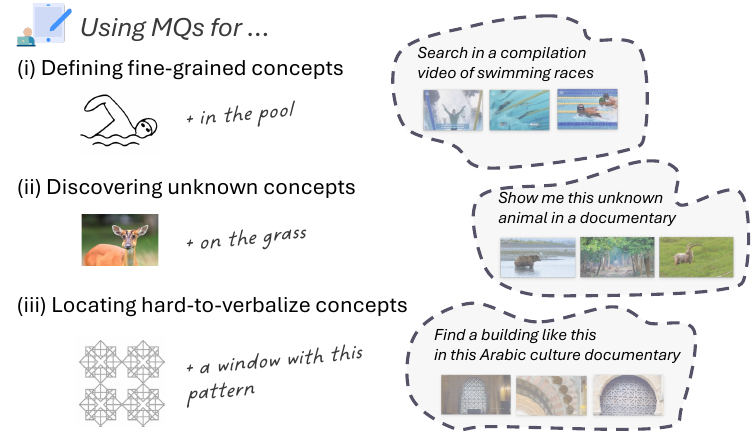}
        \caption{}
        \label{subfig:b}
    \end{subfigure}
    \caption{\textbf{Localizing Events in Videos with Semantics Queries.} Fig.~\ref{subfig:a}: So far, the community has only focused on natural language query-based video event localization as in \cite{lei2021detecting}. Our benchmark \benchmark focuses on a more general scenario: localizing events in video with multimodal queries (MQs). Fig.~\ref{subfig:b}: Localizing video events with MQs has broad applications: users often use brief, ambiguous text queries like ``swimming'' or struggle to find precise terms when it comes to unfamiliar or abstract concepts. In such cases, MQs —like scribbles or example images— can help.}
    \label{fig:teaser}
\end{figure}

Conventional video event localization encompasses a broad spectrum of related tasks in the preceding research, including
\textit{video moment retrieval}~\cite{gao2021fast,gao2021learning, liu2023survey}, \textit{highlight detection}~\cite{ badamdorj2022contrastive, lei2021detecting,moon2023query}, and \textit{video temporal grounding}~\cite{chen2021end,chen2019weakly,escorcia2019temporal,gao2021relation,hou2022cone,tan2023hierarchical, yu2024self}. 
A plethora of datasets and benchmarks~\cite{caba2015activitynet, gao2017tall, lei2021detecting,sul2024mr} has been established for exploring video event localization using Natural Language Queries (NLQs) as semantic queries.
Building on these foundations, existing models have primarily focused on this NLQ setting~\cite{anne2017localizing,chen2018temporally, chen2019localizing,chen2019semantic,chen2020learning,chen2020hierarchical,chen2019weakly,escorcia2019temporal,gao2017tall,ge2019mac, lei2021detecting,wang2023learning}.

\textit{However}, with the increasing need for human users to efficiently process massive video data online, multimodal interaction with videos is a promising scenario. 
In other words, texts should not be the only means of querying events in videos. As the saying goes, ``A picture is worth a thousand words,'' images act as a non-verbal language and convey rich semantic meaning to describe events.
For instance, as illustrated in Fig.~\ref{subfig:b}, the query ``swim" can refer to various styles of swimming, such as freestyle, butterfly, and backstroke. Using such an ambiguous query to localize fine-grained events in videos may yield imprecise results. 
As users, we often opt for writing brief, simple text queries over detailed descriptions, especially when it is hard to find the exact wording, such as unfamiliar concepts (\eg, unknown objects) or abstract ideas (\eg, aesthetic or geometric concepts).
Additionally, for illiterate users or cross-lingual use cases where texting is challenging, allowing users to search for events in videos through Multimodal Queries (MQs) like images can be beneficial.

MQs, also known as composed queries~\cite{hosseinzadeh2020composed, ventura2024covr, hummel2024egocvr, baldrati2023zero} in other contexts, offer practical benefits for video event localization. 
As illustrated in Fig.~\ref{fig:teaser}, using intuitive queries like user-drawn ``scribble images" or example images as references can enhance human-computer interaction, particularly in the scenarios described above. 
While using MQs for video event localization may seem straightforward and intuitive, several questions remain: 
(1) visual queries can introduce irrelevant or even conflicting details unrelated to the target events, and (2) visual queries align only semantically with target video events, while distribution shifts in image styles are inevitable.
How can models adapt to this more diverse and flexible MQ setting compared to the conventional NLQ-based task?

To address these questions, we propose a new task: localizing events in videos with MQs. 
We formulate an MQ consisting of a \textit{reference image}, which conveys the core semantics of the query, and a \textit{refinement text} for adjusting query details. 
This structure enables a more flexible and versatile application.
To bridge the research gap, we introduce \benchmark (\textbf{I}mage-Text \textbf{C}omposed \textbf{Q}ueries), as the first benchmark for this task, along with a new evaluation dataset, \dataset, with synthetic reference images and human-curated queries as a testbed for our task.
Considering that reference images in MQs may vary significantly from videos in terms of styles,
we define 4 reference image styles to assess performance across diverse scenarios.

Another gap to mind is that existing models designed for NLQs do not seamlessly accommodate MQs. This raises the question: how can we adapt these models for MQs?
To address this, we propose 2 Multimodal Query Adaptation (MQA) approaches, Language-Space MQA and Embedding-Space MQA, to enable preceding models as backbone models to integrate MQs.
Within these approaches, we introduce 3 training-free adaptation methods (\captioning, \summarization,\visquery) along with the Surrogate Fine-tuning on pseudo-MQs strategy, \suit, which together establish our adaptation as a SOTA baseline for video event localization using MQs.
We have selected and evaluated a broad spectrum of \modelnumber backbone models, from specialized models to Video Large Language Models (Video LLMs).

Our results demonstrate that existing models can effectively adapt to our new benchmark with MQA, establishing a solid baseline for future studies. A key insight from our findings is that, despite the potential semantic gap between MQ and NLQ, MQs remain effective for video event localization. Notably, even when MQs are minimalistic and abstract, such as scribble images, model performance is not strictly limited, envisioning new application scenarios.

Our contributions are summarized as follows:
\begin{enumerate}
    \item We introduce a new task, \textit{video event localization with MQs}, alongside a new evaluation benchmark, \benchmark, with an evaluation dataset, \dataset;  
    \item We propose 3 MQA methods and Surrogate Fine-tuning on Pseudo-MQs strategy to adapt NLQ-based backbone models;
    \item We systematically evaluate the combination of various MQA methods and \modelnumber SOTA backbone models ranging from specialized models to large-scale Video LLMs;
    \item Our comprehensive experiments demonstrate that our MQA methods offer a powerful approach for adapting existing models to \benchmark. These findings highlight the promising potential for diverse applications of MQs in video event localization. 
\end{enumerate}

%% file: sec/2_related.tex
\section{Related Work}

\begin{figure*}[htbp]
    \centering
    \includegraphics[width=.94\textwidth]{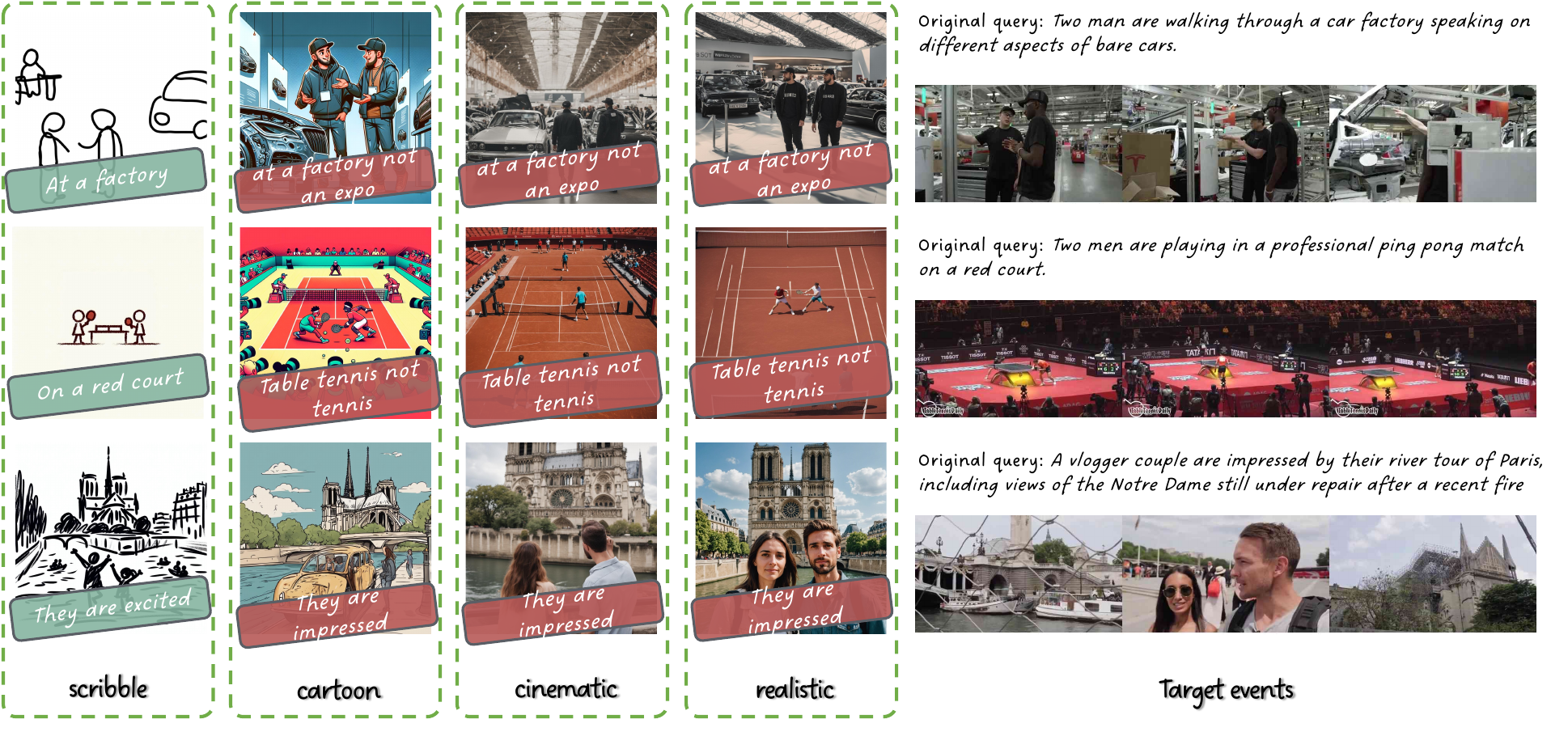}
    \caption{\textbf{Examples of \dataset.} Multimodal queries consist of a reference image and a refinement text. We consider 4 different reference image styles: scribble, cartoon, cinematic, and realistic. They describe a target event that corresponds to moments or segments in original videos and are equivalent to natural language queries in the original dataset~\cite{lei2021detecting}. Refinement texts add either {\setlength{\fboxsep}{0pt}\colorbox{comlecolor!80}{\strut complementary}} information if reference images are minimal like for scribble images, or {\setlength{\fboxsep}{0pt}\colorbox{correcolor!80}{\strut corrective}} information if reference images are more complicated.}
    \label{fig:example}
\end{figure*}

\subsection{Localizing Event in Videos with NLQs}
Query-based video temporal localization has been a long-standing research topic and is an umbrella of several related tasks. According to their scenarios and motivation, they can be further classified into several similar but slightly different tasks.
Video moment retrieval~\cite{lin2020weakly,liu2018cross, ma2020vlanet, mithun2019weakly, luo2023towards, yoon2023scanet, zala2023hierarchical, zhang2019man} aims to localize a video segment based on a textual caption query that describes events in the video.
Video temporal grounding/localization~\cite{fang2023you, hao2023fine,liu2021context,liu2021progressively,mun2020local,nam2021zero,yang2023deco,yuan2019find,zeng2020dense} with NLQs aims to determine the video segment that corresponds with textual description and usually serves downstream Question-answering task~\cite{bai2024glance, xiong2016dynamic,yu2023self,zhang2021natural} and aims to provide relevant segments in videos. 
Other similar yet less relevant tasks include video highlight detection~\cite{badamdorj2022contrastive, lei2021detecting,moon2023query,sul2024mr} and action detection; these tasks also involve localizing video segments but with an implicit query or a category-level action label.
Our benchmark steps toward localizing video events in MQs, which underlines a composed query of images and text, which are different from other works, as a semantic search for events in videos.

Regarding the methodology, a line of works is focused on NLQ-based video moment retrieval/ video temporal grounding tasks: this includes two-stage (\ie proposal-based) models~\cite{liu2021adaptive} that firstly generate moment candidates and then filter out the matched moment based on the query and one-stage (\ie proposal-free) models~\cite{chen2019localizing, rodriguez2020proposal, yuan2019find} like DETR~\cite{carion2020end}-based models have been widely employed in multiple models~\cite{jang2023knowing, lei2021detecting, moon2023query, moon2023correlation, sun2024tr, xu2023mh}.
More recent works~\cite{lin2023univtg, liu2022umt,yan2023unloc,xiao2023bridging} attempt to uniform multiple video localization tasks, including video moment retrieval and highlight detection in a single framework.
In addition, with the large-scale LLMs gaining increasing attention, temporal grounding has also been a core module in MLLMs like SeViLA~\cite{yu2023self}, InternVideo2~\cite{wang2024internvideo2}, TimeChat~\cite{ren2023timechat}, VTimeLLM~\cite{huang2024vtimellm}, \etc~\cite{zhang2023next,zhao2024videoprism}.

\vspace{-0.09cm}
\subsection{Multimodal Query for Image/Video Tasks}
Using MQs is a practical and important scenario for holistic image/video retrieval~\cite{ventura2024covr, vo2019composing,thawakar2024composed, ventura2024covr2,hummel2024egocvr, gatti2024composite, koley2024you, pal2023fashionntm, jang2024visual, saito2023pic2word, suo2024knowledge, xusentence, gu2024language,liu2023zero,jang2024spherical,wu2023few,chen2022composed}. 
Yet, it is necessary to note that video event localization with MQs \textbf{differs} from image/video retrieval tasks, which primarily involve instance-level similarity matching. Temporal localization requires dense video processing, significantly increasing the task complexity.

For video localization tasks, ~\cite{zhang2019localizing} is the first work to use image queries to localize unseen activities in videos to our knowledge. ~\cite{tian2018audio} also considers visual queries in video event localization but limits to visual-audio data.
More recently, ~\cite{goyal2023minotaur} proposes to ground videos spatio-temporally using images or texts, although their queries are still limited to object or action levels. To the best of our knowledge, our work is the first to attempt localizing events in videos using multimodal semantic queries.

%% file: sec/3_benchmark.tex
\section{Video Event Localization with Multimodal Queries: A Testbed}
In the following section, we will elaborate on the definition of our new task, the benchmark \benchmark, and \dataset.

\subsection{Task Definition}
We define a multimodal query \qm as consisting of a reference image \vref accompanied by a refinement text \tmod for minor adjustments for localizing a target event that corresponds to the query semantically. The reference image captures the key semantics of the target event, while the refinement text provides extra information that can be either \textit{complementary} or \textit{corrective}. This enables multimodal queries to be more adaptable to real-world applications. 

Given the query \qm, the model predicts \textit{all} the relevant segments or moments $\left[\tau_{start}, \tau_{end}\right]$.
We employ recall and mean Average Precision as the evaluation metrics for this task as NLQ-based localization.

\begin{figure*}[htbp]
    \centering
    \includegraphics[width=0.99\textwidth]{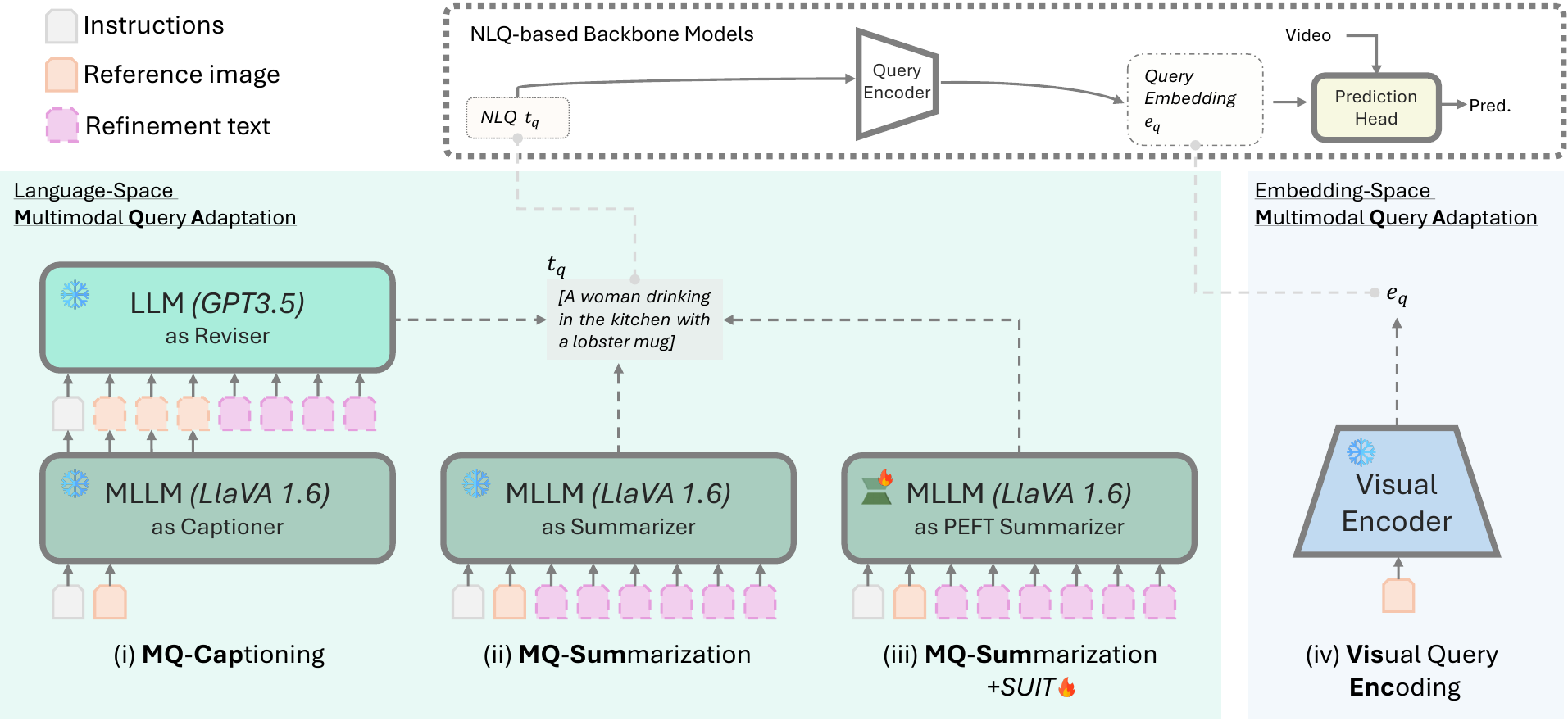}
    \caption{\textbf{Multimodal Query Adaptation (MQA).} We propose 3 MQA methods to bridge the current gap between natural language query-based models and our multimodal query-based benchmark: \captioning, \summarization, and \visquery and \sumsuit enhanced by Surrogate Fine-tuning on pseudo-MQs (\sumsuit) strategy, to adapt MQs to the conventional NLQ-based backbones.}
    \label{fig:adapt}
\end{figure*}

\vspace{0.2cm}
\noindent
\textbf{Reference Image}
Reference images \vref visually describe the semantics of an event in a video. 
They can be simple scribble images with minimal strokes that describe an event succinctly, effectively summarizing an event for non-verbal semantic queries in video localization or more detailed images that depict semantically relevant scenes in a video.  
As illustrated in Fig.~\ref{fig:example}, reference images describe semantically similar scenes yet might vary in details as target videos.
In practice, visual queries can differ in style, which may impact model performance. Therefore, we explore multiple reference image styles, as detailed in the subsequent section, to assess whether the model maintains consistent performance across various styles.

\vspace{0.2cm}
\noindent
\textbf{Refinement Texts}
Refinement texts refer to simple phrases to \textit{complement} or \textit{correct} descriptions that are either missing or contradictory in the reference images. This is particularly practical in real-world applications, as reference images often do not semantically align perfectly with the target video event. 
We identify 5 different types of refinement texts that can be applied to various aspects of the reference image semantics: ``object'', ``action'', ``relation'', ``attribute'', ``environment'', and ``others'' as shown in Fig.~\ref{fig:dist} in Appx.~\ref{sec:stats}. This categorization is designed for elements of a semantic scene graph~\cite{ji2020action} and we borrowed it to summarize different semantic elements of the multimodal queries.

\subsection{Dataset Construction}\label{sec:data}
We introduce our new evaluation dataset, \dataset, as a testbed for \benchmark. This dataset is built upon the validation set of QVHighlights~\cite{lei2021detecting}, a popular NLQ-based video localization dataset.
For each original query in QVHighlights, we construct multimodal semantic queries that incorporate reference images paired with refinement texts. 
Considering the reference image style distribution discussed earlier, \dataset features 4 variants based on different image styles. 
Detailed statistics can be found in Appx.~\ref{sec:data-app}.

\vspace{0.2cm}
\noindent
\textbf{Reference Image Generation} 
We generate reference images based on the original natural language queries and refinement texts using a suite of state-of-the-art Text-to-Image (T2I) models, including DALL-E-2\footnote{\url{https://openai.com/index/dall-e-2/}} and Stable Diffusion\footnote{\url{https://stability.ai/stable-image}}. For the reference image styles mentioned earlier, we select 4 representative styles: \scribble, \cartoon, \cinematic, and \realistic. These styles effectively capture a variety of real-world scenarios such as user inputs, book illustrations, television shows, and actual photographs, where images are often used as queries.

\vspace{0.2cm}
\noindent
\textbf{Data Annotation and Preprocessing} 
We emphasize the meticulous crowd-sourced data curation and annotation effort applied to QVHighlights for 2 main reasons: (1) To introduce refinement texts, we purposefully modify the original semantics of text queries in QVHighlights to generate queries that are similar yet subtly different; (2) Given that the original queries in QVHighlights can be too simple and ambiguous to generate reasonable reference images, we add necessary annotations to ensure that the generated image queries are more relevant to the original video semantics. 
We employed human annotators to annotate and modify the natural language queries. Each query is annotated and reviewed by different annotators to ensure consistency. 
Further details can be found in the Appx.~\ref{sec:data-app}.

\section{Adapting Multimodal Query}
To explore the performance of preceding NLQ-based video localization methods on \benchmark, we propose 2 Multimodal Query Adaptation (MQA) (in Sec.~\ref{sec:mqa}) strategies to bridge the gap between natural language queries (NLQs) and multimodal queries (MQs): Language-Space MQA and Embedding-Space MQA. Among them, we propose 3 training-free methods that adapt MQs to NLQs and a parameter-efficient fine-tuned (PEFT) method tailored for MQA task with a novel Surrogate Fine-tuning strategy (in Sec.~\ref{sec:suit}). In total, we have benchmarked \modelnumber video event localization models (in Sec.~\ref{sec:baseline}) for a thorough evaluation.



\subsection{Multimodal Query Adaptation}\label{sec:mqa}
In the conventional paradigm, input NLQs $t_q$ are embedded in a high-dimensional space as query embeddings $e_q$. A common practice is leveraging CLIP~\cite{radford2021learning} text encoder as the query encoder shown in Tab.~\ref{tab:model_comp} in Appx.~\ref{sec:compare}.

To align the MQs with pre-trained NLQs, we categorize MQA by different adaptation stages: Language-Space MQA, where MQs are transcribed to NLQs, and Embedding-Space MQA, where MQs are directly encoded as query embeddings, without transcription, as illustrated in Fig.~\ref{fig:adapt}.

For Language-Space MQA, we first propose 2 training-free methods, MQ-Captioning (\captioning) and MQ-Summarization (\summarization), to leverage the power of MLLMs.
\captioning uses MLLMs as a captioner to caption reference images and LLMs as a modifier to integrate refinement texts. In contrast, \summarization utilizes MLLMs to directly summarize reference images and refinement texts in one step. Generated texts $t_{q}$ can be seamlessly used by existing models.

For Embedding-Space MQA, we propose Visual Query Encoding (\visquery) using only reference images to embed the reference images as query embeddings $e_{q}$. This is based on the precondition that all selected models employ a dual-stream encoder that embeds image-text pairs in a joint embedding space.

Nevertheless, such methods still confront some performance issues (discussed in Sec.~\ref{sec:ana}), including i) different prompt selection causes unstable performance; ii) MLLMs tend to generate overly long and less task-specific outputs, which lead to NLQ distribution shift that backbone models rely on and harm the model performance. Therefore, we also propose a MLLM strategy for MQA, which is called Surrogate Fine-tuning on pseudo-MQs for MQA.

\begin{table*}[htbp]
\scriptsize
\setlength{\tabcolsep}{8pt}
\centering
\vspace{-0.4cm}
\begin{adjustbox}{center}
\begin{NiceTabular}{llcccccccc}[colortbl-like]
\CodeBefore
\rowcolor{gray!15}{3-8,12-17,21-23,30-32,37-42}
\rowcolor{white}{9-11}
\columncolor{white}{1}
\Body
\toprule
\multirow{2}{*}{\textbf{}} & \multirow{2}{*}{\textbf{Model}}
& \multicolumn{2}{c}{\textbf{\texttt{scribble}}} & 
\multicolumn{2}{c}{\textbf{\texttt{cartoon}}} &
\multicolumn{2}{c}{\textbf{\texttt{cinematic}}}
& \multicolumn{2}{c}{\textbf{\texttt{realistic}}} \\
& & R1@0.5 & R1@0.7 & R1@0.5 & R1@0.7 & R1@0.5 & R1@0.7 & R1@0.5 & R1@0.7 \\
\midrule
\multirow{9}{*}{\rotatebox{90}{\visquery}} &
Moment-DETR (2021) & 12.55 & 5.69 & 13.38 & 6.59 & 14.36 & 6.01 & 14.88 & 6.53 \\
& QD-DETR (2023) & 15.91 & 9.12 & 14.88 & 8.62 & 13.90 & 8.49 & 14.62 & 8.36 \\
& QD-DETR$\dag$ (2023) & 15.65 & 10.03 & 12.60 & 6.79 & 12.34 & 6.72 & 12.34 & 7.44 \\
& EaTR (2023) & 19.86 & \textit{13.00} & 19.91 & 12.99 & 21.15 & \textit{13.45} & 21.48 & 13.38 \\
& CG-DETR (2023) & \textit{22.90} & \textit{13.00} & \textit{24.93} & 13.58 & \textit{23.24} & 13.12 & \textit{24.74} & \textit{14.23} \\
& TR-DETR (2024) & 17.92 & 11.19 & 17.36 & 11.10 & 15.14 & 9.86 & 15.60 & 9.53 \\
& UMT$\dag$ (2022) & 5.43 & 2.85 & 4.77 & 2.09 & 5.22 & 2.35 & 4.57 & 2.42 \\
& UniVTG (2023) & 21.93 & 13.00 & 23.89 & \textit{13.64} & 22.78 & 13.19 & 22.52 & 12.79 \\
& UVCOM (2023) & 17.08 & 9.77 & 16.78 & 10.97 & 17.36 & 11.68 & 17.10 & 11.23 \\
\midrule
\multirow{10}{*}{\rotatebox{90}{\captioning}}
& Moment-DETR (2021) & 44.83 \rel{± 2.7} & 27.97 \rel{± 2.2} & 46.02 \rel{± 1.5} & 29.36 \rel{± 0.9} & 46.89 \rel{± 0.7} & 30.35 \rel{± 1.2} & 47.16 \rel{± 1.5} & 30.53 \rel{± 0.8} \\
& QD-DETR (2023) & 48.92 \rel{± 4.1} & 33.57 \rel{± 3.3} & 52.87 \rel{± 0.8} & 36.01 \rel{± 1.3} & 54.01 \rel{± 0.7} & 37.29 \rel{± 0.5} & 53.07 \rel{± 0.8} & 37.53 \rel{± 1.1} \\
& QD-DETR$\dag$ (2023) & 50.15 \rel{± 4.6} & 34.67 \rel{± 3.9} & 53.53 \rel{± 1.3} & 38.30 \rel{± 1.2} & 53.37 \rel{± 0.6} & 37.93 \rel{± 0.5} & 53.39 \rel{± 1.0} & 38.47 \rel{± 0.8} \\
& EaTR (2023) & 49.20 \rel{± 3.2} & 34.82 \rel{± 3.5} & 50.50 \rel{± 0.6} & 35.27 \rel{± 0.7} & 51.76 \rel{± 0.5} & 36.92 \rel{± 0.7} & 52.33 \rel{± 0.5} & 37.01 \rel{± 0.3} \\
& CG-DETR (2023) & 50.65 \rel{± 3.5} & 36.37 \rel{± 2.9} & \textit{56.26 \rel{± 0.7}} & \textit{40.82 \rel{± 0.7}} & 54.53 \rel{± 0.9} & 39.32 \rel{± 0.8} & 56.72 \rel{± 0.7} & 41.79 \rel{± 1.2} \\
& TR-DETR (2024) & \textit{50.99 \rel{± 3.3}} & 35.55 \rel{± 3.7} & 55.37 \rel{± 1.0} & 39.92 \rel{± 2.0} & \textit{56.03 \rel{± 1.0}} & 40.69 \rel{± 0.9} & \textit{56.94 \rel{± 0.5}} & 41.99 \rel{± 0.3} \\
& UMT$\dag$ (2022) & 44.76 \rel{± 3.5} & 29.41 \rel{± 3.0} & 48.15 \rel{± 1.7} & 32.18 \rel{± 1.6} & 49.96 \rel{± 0.9} & 33.90 \rel{± 0.9} & 48.83 \rel{± 1.0} & 34.09 \rel{± 1.2} \\
& UniVTG (2023) & 47.50 \rel{± 3.1} & 31.58 \rel{± 3.0} & 49.50 \rel{± 0.8} & 33.09 \rel{± 1.1} & 50.98 \rel{± 0.2} & 33.36 \rel{± 0.6} & 51.42 \rel{± 1.1} & \textit{43.75 \rel{± 0.2}} \\
& UVCOM (2023) & \textit{50.99 \rel{± 3.6}} & \textit{37.36 \rel{± 3.1}} & 54.39 \rel{± 0.5} & 40.06 \rel{± 1.0} & 55.88 \rel{± 0.7} & \textit{40.88 \rel{± 0.5}} & 54.92 \rel{± 0.9} & 41.08 \rel{± 0.9} \\
& SeViLA (2023) & 17.37 \rel{± 1.3}& 10.56 \rel{± 0.8}& 22.72 \rel{± 0.8}& 15.31 \rel{± 0.7} & 25.94 \rel{± 0.1} & 16.99 \rel{± 0.3}& 26.83 \rel{± 0.8} & 16.83 \rel{± 0.6}\\
& TimeChat (2024) & 6.63 \rel{± 0.8} & 3.07 \rel{± 0.7} & 8.24 \rel{± 1.0} & 3.62 \rel{± 0.8} & 8.15 \rel{± 0.6} & 3.15 \rel{± 0.4} & 7.70 \rel{± 0.5} & 3.17 \rel{± 0.5}\\
& VTimeLLM (2024) & 16.24 \rel{± 0.9} &	6.98 \rel{0.4} & 19.49 \rel{± 0.4} & 7.86 \rel{± 0.2}	& 20.9 \rel{± 0.4} & 8.64 \rel{± 0.4} & 20.75 \rel{± 0.5} & 8.67 \rel{± 0.2} \\
\midrule
\multirow{10}{*}{\rotatebox{90}{\summarization}}
& Moment-DETR (2021) & 42.00 \rel{± 3.3} & 25.14 \rel{± 3.0} & 44.56 \rel{± 2.4} & 27.24 \rel{± 2.1} & 43.73 \rel{± 2.0} & 27.00 \rel{± 1.8} & 44.34 \rel{± 2.6} & 27.74 \rel{± 2.0} \\
& QD-DETR (2023) & 45.56 \rel{± 3.3} & 30.44 \rel{± 3.0} & 49.09 \rel{± 3.8} & 33.64 \rel{± 3.2} & 48.89 \rel{± 3.5} & 32.66 \rel{± 3.1} & 47.83 \rel{± 4.1} & 32.86 \rel{± 3.8} \\
& QD-DETR$\dag$ (2023) & 46.57 \rel{± 3.8} & 32.52 \rel{± 3.6} & 49.30 \rel{± 4.3} & 34.12 \rel{± 4.2} & 48.83 \rel{± 3.2} & 34.16 \rel{± 3.4} & 49.13 \rel{± 4.4} & 33.83 \rel{± 3.1} \\
& EaTR (2023) & 45.79 \rel{± 3.0} & 32.67 \rel{± 2.9} & 48.45 \rel{± 2.9} & 32.96 \rel{± 2.7} & 48.24 \rel{± 3.8} & 33.35 \rel{± 3.5} & 48.69 \rel{± 3.7} & 33.85 \rel{± 2.5} \\
& CG-DETR (2023) & \textit{47.07 \rel{± 4.2}} & 33.14 \rel{± 4.1} & 51.46 \rel{± 3.1} & 36.49 \rel{± 2.7} & 50.59 \rel{± 3.4} & 36.08 \rel{± 3.6} & 51.91 \rel{± 3.5} & 36.58 \rel{± 2.4} \\
& TR-DETR (2024) & 46.44 \rel{± 4.4} & 33.23 \rel{± 3.8} & 51.35 \rel{± 3.2} & 36.14 \rel{± 2.3} & \textit{51.92 \rel{± 3.8}} & 36.29 \rel{± 3.7} & \textit{52.87 \rel{± 4.0}} & \textit{36.77 \rel{± 3.4}} \\
& UMT$\dag$ (2022) & 43.88 \rel{± 3.4} & 29.28 \rel{± 1.9} & 45.39 \rel{± 2.8} & 29.98 \rel{± 2.4} & 45.37 \rel{± 2.3} & 30.01 \rel{± 2.2} & 46.35 \rel{± 2.0} & 30.27 \rel{± 1.0} \\
& UniVTG (2023) & 44.98 \rel{± 3.3} & 27.99 \rel{± 2.7} & 46.19 \rel{± 3.5} & 30.37 \rel{± 2.4} & 47.22 \rel{± 3.3} & 29.90 \rel{± 2.5} & 50.39 \rel{± 3.3} & 30.33 \rel{± 2.4} \\
& UVCOM (2023) & 46.62 \rel{± 3.8} & \textit{33.40 \rel{± 3.4}} & \textit{51.48 \rel{± 4.1}} & \textit{36.92 \rel{± 3.7}} & 50.91 \rel{± 5.3} & \textit{36.58 \rel{± 4.5}} & 51.18 \rel{± 3.7} & 36.23 \rel{± 3.4} \\
& SeViLA (2023) &17.89 \rel{± 1.9}& 10.65 \rel{± 1.5} & 27.47 \rel{± 3.5}& 16.98 \rel{± 1.9}& 27.76 \rel{± 2.5} & 17.77 \rel{± 1.5}& 28.61 \rel{± 3.3} &
17.30 \rel{± 2.0}\\
& TimeChat (2024) & 6.58 \rel{± 0.1} & 2.76 \rel{± 0.5}& 7.38 \rel{± 1.1}& 3.39 \rel{± 0.8}& 7.51 \rel{± 0.9} & 3.63 \rel{± 0.8} & 5.73 \rel{± 1.2} & 4.49 \rel{± 3.3}\\
& VTimeLLM (2024) & 16.95 \rel{± 1.4}  & 7.40 \rel{± 0.1}  & 19.19 \rel{± 0.8} & 7.8 \rel{± 0.3} &	20.23 \rel{± 0.4} &	8.29 \rel{± 0.3} &	20.53 \rel{± 1.5} &	8.11 \rel{± 0.5}  \\
\midrule
\multirow{10}{*}{\rotatebox{90}{\summarization}} 
&  \underline{\textit{\textbf{+ SUIT}}}\\
& Moment-DETR (2021) &  48.59 \rel{± 0.9} & 31.85 \rel{± 0.7} & 48.27 \rel{± 0.6} & 31.31 \rel{± 0.4} & 47.58 \rel{± 0.5} & 31.52 \rel{± 0.5} & 47.25 \rel{± 0.2} & 30.83 \rel{± 0.6} \\
& QD-DETR (2023) & 55.27 \rel{± 0.5} & 39.86 \rel{± 0.4} & 53.45 \rel{± 0.6} & 37.94 \rel{± 0.3} & 53.36 \rel{± 0.3} & 38.39 \rel{± 0.6} & 53.79 \rel{± 0.5} & 38.92 \rel{± 0.1}\\
& QD-DETR\dag (2023) & 55.20 \rel{± 0.5} & 39.82 \rel{± 0.7} & 54.60 \rel{± 0.4} & 40.44 \rel{± 0.6} & 54.28 \rel{± 0.4} & 40.31 \rel{± 0.6} & 53.52 \rel{± 0.8} &
38.97 \rel{± 0.1} \\
& EaTR (2023) & 53.63 \rel{± 0.8} & 39.23 \rel{± 0.5} & 50.63 \rel{± 0.4} & 37.40 \rel{± 0.6} & 51.67 \rel{± 0.5} & 38.50 \rel{± 0.4} & 50.78 \rel{± 0.4} & 37.19 \rel{± 0.5} \\
& CG-DETR (2023) & 55.83 \rel{± 0.6} & 41.41 \rel{± 0.3} & 55.42 \rel{± 0.8} & 39.88 \rel{± 0.6} & 56.37 \rel{± 0.8} & 41.14 \rel{± 0.6} & 55.47 \rel{± 0.9} & 40.17 \rel{± 0.5} \\
& TR-DETR (2024) & \textbf{\textit{58.85}} \rel{± 0.4} & \textbf{\textit{43.08}} \rel{± 0.4} & \textbf{\textit{57.19}} \rel{± 0.2} & \textbf{\textit{41.31}} \rel{± 0.4} & \textit{\textbf{57.35}} \rel{± 0.5}  & \textit{\textbf{41.92}} \rel{± 0.9} & \textbf{\textit{57.39}} \rel{± 0.4} & \textbf{\textit{42.64}} \rel{± 0.3}\\
& UMT\dag (2022) & 49.71 \rel{± 0.3} & 35.10 \rel{± 0.3} & 50.01 \rel{± 0.8} & 35.16 \rel{± 0.6} & 50.25 \rel{± 0.6} & 35.18 \rel{± 0.5}  & 49.85 \rel{± 0.4} & 34.60 \rel{± 0.7} \\
& UniVTG (2023) & 51.26 \rel{± 0.4} & 34.07 \rel{± 0.7} & 49.36 \rel{± 0.3} & 33.24 \rel{± 0.5} & 51.0 \rel{± 0.5} & 34.4 \rel{± 0.7} & 50.65 \rel{± 0.6} & 33.48 \rel{± 0.6} \\
& UVCOM (2023) & 55.33 \rel{± 0.4} & 42.03 \rel{± 0.7} & 55.48 \rel{± 0.2} & 41.66 \rel{± 0.1} & 55.43 \rel{± 0.4}& 41.88 \rel{± 0.4} & 54.43 \rel{± 0.4} & 41.30 \rel{± 0.3} \\

\bottomrule
\end{NiceTabular}
\end{adjustbox}
\caption{\textbf{Model performance (Recall) on \benchmark.} We highlight the best score in \textit{italic} for each adaptation method and the overall best scores in \textbf{bold}.
For \captioning and \summarization, we report the standard deviation of 3 runs with different prompts, and for \sumsuit, we report the average performance with different seeds in training. $\dag$ uses extra audio modality.}
\label{tab:main-results-recall}
\end{table*}

\subsection{SUIT: Surrogate Fine-tuning on Pseudo-MQs}\label{sec:suit}
Fine-tuning MLLMs on the task of summarizing MQs could counteract the impact of different prompt selections and mitigate the distribution shift between original NLQs and generated NLQs.
\textit{However}, an underlying challenge for fine-tuning lies in the lack of training data for MQ-based localization. Compared to establishing an evaluation testbed, the larger-scale training data is more time and labor-intensive. 
Besides, synthetic training data could pose risks of overfitting on generation bias and artifacts in the model, which is supposed to be avoided.

To overcome this challenge, we propose a novel strategy, \textbf{SU}rrogate F\textbf{I}ne-\textbf{T}uning (SUIT) on pseudo-MQs, to alleviate the training data issue.

\begin{figure}
    \centering
    \includegraphics[width=1.\linewidth]{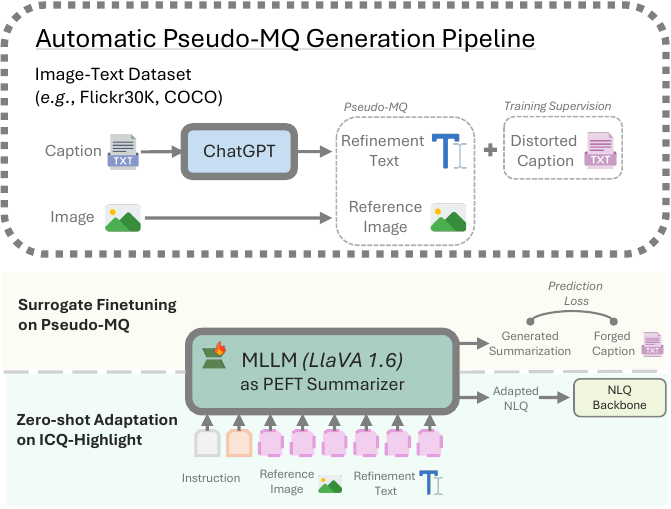}
    \caption{\textbf{Surrogate Fine-tuning on pseudo-MQs (SUIT).} for MQ-Sum. To solve the issue of lacking training data, we propose an automatic pseudo-MQ generation pipeline to construct a ``surrogate" dataset for fine-tuning \summarization.}
    \label{fig:suit}
\end{figure}

As illustrated in Fig.~\ref{fig:suit}, SUIT consists of 2 steps:

\vspace{0.2cm}
\noindent
\textbf{Pseudo-MQ Generation Pipeline}
To deal with the insufficient training data problem, we propose leveraging the abundant image-text datasets like Flickr30K~\cite{jia2015guiding} and COCO~\cite{lin2014microsoft} to generate pseudo-MQs. We automate this generation process by leveraging GPT3.5 to convert each caption in the datasets to a pair of a ``forged" caption and a refinement text that reflects the forge. As a result, the original image and the refinement text constitute a pseudo-MQ that is equivalent to a forged caption semantically.

\vspace{0.2cm}
\noindent
\textbf{Surrogate Fine-tuning on Psuedo-MQs}
We further utilize generated pseudo-MQs as inputs and instruct MLLMs to generate a summarization as \summarization. Distorted captions are used as supervision to fine-tune the model with the next-token prediction loss and the PEFT approach as a surrogate training task.
Then, we can transfer the fine-tuned MLLMs to our \dataset dataset for evaluation.

\subsection{Backbone Model Selection}\label{sec:baseline}
We have selected and benchmarked \modelnumber models specifically designed for video event localization with NLQs. 
Particularly, we categorize the selected models as follows and compare the models in different dimensions in the Appendix: (1) \textit{Specialized models} use natural language as a semantic query and are targeted at video moment retrieval tasks. We have selected a series of these models including Moment-DETR\cite{lei2021detecting}, QD-DETR\cite{moon2023query}, EaTR\cite{jang2023knowing}, CG-DETR\cite{moon2023correlation}, and TR-DETR\cite{sun2024tr};
(2) \textit{Unified frameworks} are aimed to solve multiple video localization tasks within one model, such as moment retrieval, highlight detection, and video summarization. We have selected UMT\cite{liu2022umt}, UniVTG\cite{lin2023univtg}, and UVCOM\cite{xiao2023bridging} as strong baselines;
(3) \textit{LLM-based Models} features the power of Large Language Models, which prove to be a powerful and general head for varied video tasks. We have selected SeViLA~\cite{yu2023self}, TimeChat~\cite{ren2023timechat}, and VTimeLLM~\cite{huang2024vtimellm} as representatives of LLM-based models. 
We apply different MQA methods on top of the pre-trained model checkpoints that have been fine-tuned on the original QVHighlights dataset.

\begin{figure}[htbp]
    \centering
        \centering
        \centering
\includegraphics[width=.82\linewidth]{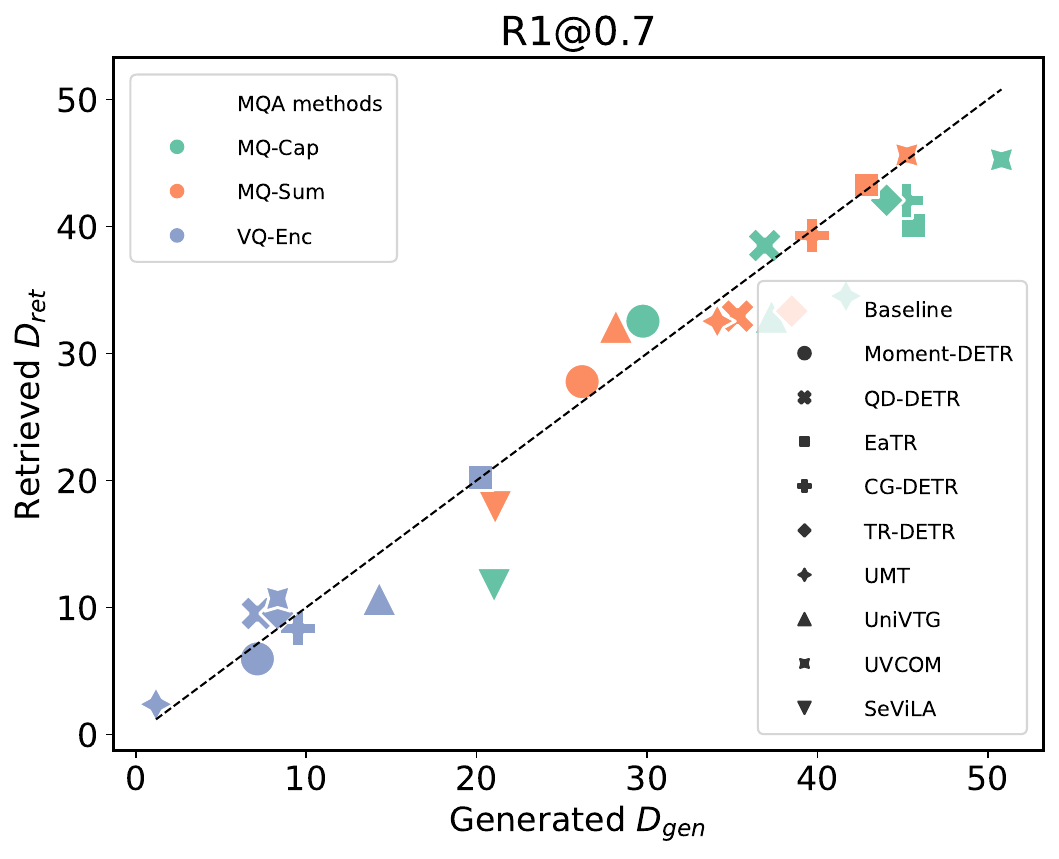}
    \vspace{-0.3cm}
    \caption{\textbf{Controlled Experiment.} We plot the model performance (R1@0.7) on 2 subsets $D_{ret}$ and $D_{gen}$. We use the dashed line to indicate the same performance on both datasets.}
    \label{fig:control}
\end{figure}

%% file: sec/4_experiment.tex
\section{Experiments and Analysis}\label{sec:ana}
In this section, we attempt to answer the following questions:
(1) Can and how well MQs effectively localize events in videos?
(2) Can varied styles of reference images and refinement texts impact the results?

\subsection{Experimental Setup}
\vspace{0.2cm} 
\noindent 
\textbf{Implementation} 
We employ LLaVA-mistral-1.6~\cite{liu2023improvedllava, liu2023llava} as a strong MLLM in \captioning, \summarization (with and without SUIT) and GPT-3.5 as a reviser in our \captioning adaptation.
We believe that the performance of these models is representative of the SOTA capabilities of MLLMs and is fairly compared across different MQA methods.
For \visquery, we utilize the corresponding CLIP Visual Encoder, as all models typically employ the CLIP Text Encoder for text query encoding. In this adaptation method, we omit refinement texts and only use the reference image.
In \sumsuit, we construct our pseudo-MQs with $\num{89420}$ training data from Flickr30K and COCO and implement LoRA~\cite{hu2021lora} as a common PEFT method with rank $32$, alpha $64$, and a learning rate of $2\times10^{-4}$ on language model of LlaVA. More implementation details about datasets and training can be found in the Appx.~\ref{sec:append_imple}.

\vspace{0.2cm} 
\noindent 
\textbf{Evaluation Metrics}
We evaluate models on our new testbed \dataset.
For evaluation, we report both Recall R@1 with IoU thresholds $0.5$ and $0.7$, mean Average Precision with IoU threshold 0.5 and the average over multiple IoU thresholds [0.5:0.05:0.95] as standard metrics for video moment retrieval and localization~\cite{lei2021detecting, yu2023self}, where IoU (Intersection over Union) thresholds determine if a predicted temporal window is positive.

\begin{table*}[htbp]
\fontsize{6}{8}\selectfont
\scriptsize
\setlength{\tabcolsep}{8pt}
\centering
\vspace{-0.4cm}
\begin{tabular}{lcccccccc}
\toprule
\multirow{2}{*}{\textbf{Model}} & \multicolumn{2}{c}{\textbf{\scribble}} & \multicolumn{2}{c}{\textbf{\cartoon}} & \multicolumn{2}{c}{\textbf{\cinematic}}  & \multicolumn{2}{c}{\textbf{\realistic}}\\
 & R1@0.5 & R1@0.7 & R1@0.5 & R1@0.7
 & R1@0.5 & R1@0.7 & R1@0.5 & R1@0.7\\
\midrule
Moment-DETR & 45.15 \tiny{(-2.7$\%$)} & 28.72 \tiny{(-3.3$\%$)} & 43.60 \tiny{(-7.1$\%$)} & 27.94 \tiny{(-5.8$\%$)} & 44.06 \tiny{(-7.3$\%$)} & 29.70 \tiny{(-2.8$\%$)} & 44.06 \tiny{(-9.3$\%$)} &28.98 \tiny{(-6.5$\%$)}\\
QD-DETR & 49.81 \tiny{(-4.0$\%$)} & 33.70 \tiny{(-5.4$\%$)} & 49.87 \tiny{(-6.6$\%$)} & 34.33 \tiny{(-6.3$\%$)} & 49.67 \tiny{(-9.3$\%$)} & 34.73 \tiny{(-8.1$\%$)} & 50.52 \tiny{(-5.7$\%$)} & 35.25 \tiny{(-7.4$\%$)}\\
QD-DETR$\dag$ & 51.29 \tiny{(-3.9$\%$)} & 36.03 \tiny{(-3.8$\%$)} & 48.69 \tiny{(-10.8$\%$)}  & 33.88 \tiny{(-13.4$\%$)} & 49.48 \tiny{(-8.5$\%$)} & 34.99 \tiny{(-9.0$\%$)} & 49.93 \tiny{(-7.5$\%$)} & 35.05 \tiny{(-10.4$\%$)}\\
EaTR & \cellcolor{white}52.01 \tiny{(+0.5$\%$)} & \cellcolor{white}37.77 \tiny{(+1.2$\%$)} & 47.45 \tiny{(-6.7$\%$)} & 33.09 \tiny{(-8.0$\%$)} & 48.56 \tiny{(-7.0)} & 34.33 \tiny{(-5.1)} & 49.61 \tiny{(-6.1$\%$)} & 35.64 \tiny{(-3.0$\%$)}\\
CG-DETR & 51.42 \tiny{(-4.0$\%$)} & 37.84 \tiny{(-1.7$\%$)} & 49.35 \tiny{(-13.0$\%$)} & 35.90 \tiny{(-13.4$\%$)} & 48.89 \tiny{(-10.3)} & 34.79 \tiny{(-11.3)} & 51.04 \tiny{(-10.5$\%$)} & 36.55 \tiny{(-14.0$\%$)}\\
TR-DETR & 52.01 \tiny{(-2.4$\%$)} & 37.19 \tiny{(-2.9$\%$)} & 51.04 \tiny{(-9.2$\%$)} & 36.62 \tiny{(-11.2$\%$)} & 50.00 \tiny{(-11.8)} & 36.03 \tiny{(-12.5)} & 52.28 \tiny{(-8.8$\%$)} & 37.53 \tiny{(-10.6$\%$)}\\
UMT$\dag$ & 46.25 \tiny{(-3.0$\%$)} & 31.57 \tiny{(-1.0$\%$)} & 45.82 \tiny{(-6.9$\%$)} & 30.61 \tiny{(-7.1$\%$)} & 46.34 \tiny{(-8.6$\%$)} & 29.96 \tiny{(-13.7$\%$)} & 46.08 \tiny{(-6.2$\%$)} & 31.85 \tiny{(-7.1$\%$)}\\
UniVTG & 47.87 \tiny{(-3.8$\%$)} & 33.76 \tiny{(-2.2$\%$)} & 45.56 \tiny{(-9.4$\%$)} & 29.24 \tiny{(-11.5$\%$)} & 45.43 \tiny{(-11.2$\%$)} & 29.05 \tiny{(-13.9$\%$)} & 46.80 \tiny{(-9.3$\%$)} & 30.42 \tiny{(-12.4$\%$)}\\
UVCOM & 52.26 \tiny{(-1.7$\%$)} & \cellcolor{white}39.39 \tiny{(+1.0$\%$)} & 51.50 \tiny{(-6.1$\%$)} & 37.99 \tiny{(-6.6$\%$)} & 50.98 \tiny{(-9.4$\%$)} & 36.75 \tiny{(-11.3$\%$)} & 51.70 \tiny{(-7.6$\%$)} & 37.53 \tiny{(-10.5$\%$)}\\
SeViLA & 13.15 \tiny{(-30.3$\%$)} & 8.06 \tiny{(-29.3$\%$)} & 11.89 \tiny{(-49.8$\%$)}& 6.89 \tiny{(-57.0$\%$)}  & 13.26 \tiny{(-49.0$\%$)}& 8.32 \tiny{(-51.5$\%$)} & 13.65 \tiny{(-49.1$\%$)} & 8.22 \tiny{(-51.1$\%$)}\\
\bottomrule
\end{tabular}
\caption{\textbf{Model performance without refinement texts.} We employ \captioning for methods without considering refinement texts. The performance drop highlighted in the parenthesis indicates that refinement texts in \dataset can help refine the semantics of the reference images and localize the events better.}
\label{tab:wo_refinement}
\end{table*}

\begin{figure}[hb]
    \centering     
\vspace{-0.4cm}
    \includegraphics[width=0.82\linewidth]{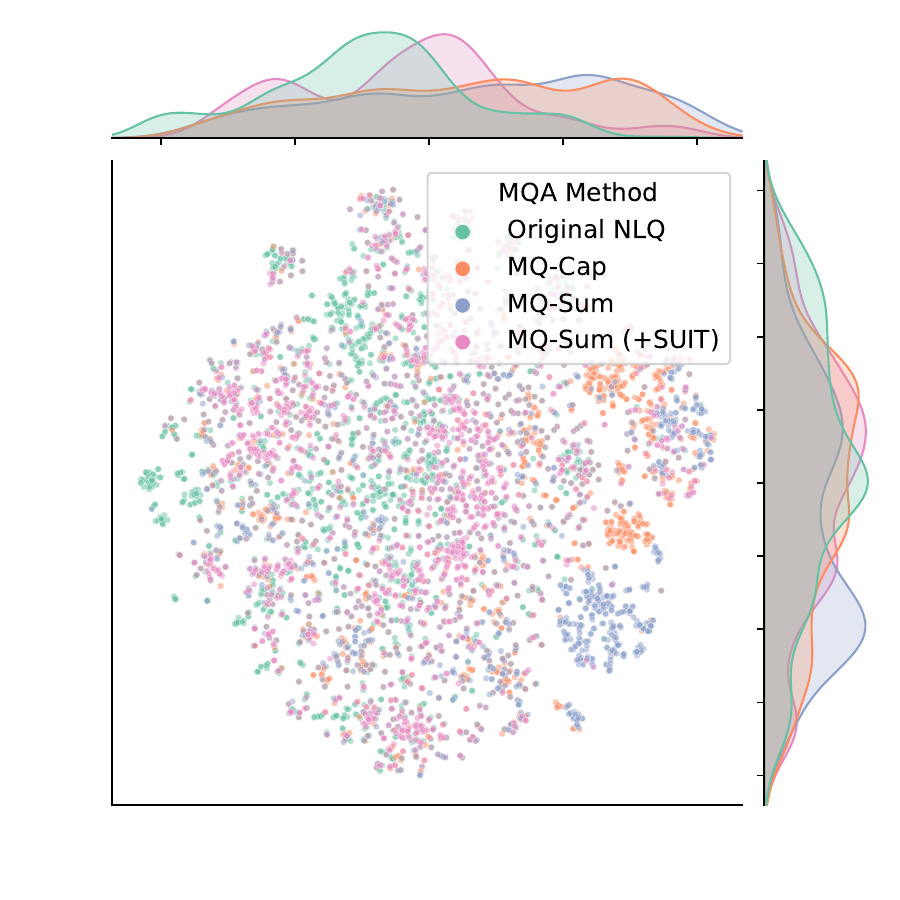}
    \vspace{-0.7cm}
    \caption{\textbf{t-SNE Visualization of Queries after Language-Space Multimodal Query Adaptation.} Original NLQs have similar distributions with closer modes as \sumsuit other than the other two training-free methods, which shows that finetuned MLLM can generate closer queries to original NLQs.}
    \label{fig:vis}
\end{figure}

\subsection{Results \& Analysis}\label{sec:results}
We present the pairwise performance of \modelnumber models combined with 4 adaptation methods on \benchmark in Tab.~\ref{tab:main-results-recall} and Tab.~\ref{tab:main-results-map} in Appx.~\ref{sec:moreresults}. For \captioning and \summarization methods, we have conducted multiple runs with different prompts and reported the average performance and standard deviation.

\vspace{0.2cm} 
\noindent 
\textbf{How do Video Event Localization with MQs work on different image styles?}
Firstly, we aim to draw a key conclusion from the results.
We find all adaptation methods perform consistently across different styles and therefore suggest that they could understand the MQs well, particularly for styles including \cartoon, \cinematic, and \realistic; the model performance is close to each other. For \scribble, all models show marginally worse performance, and even both \captioning and \summarization methods have a more significant standard deviation, which reflects that it is heavily influenced by the prompts. This can be explained by the fact that \scribble images are more minimal and abstract in semantics and more challenging to interpret.
\textit{Surprisingly}, in spite of being more abstract and simpler, the model performance on \scribble reference images is close to other reference image styles. This demonstrates the potential of using \scribble as MQs in real-world video event localization applications like video search.

\vspace{0.2cm} \noindent \textbf{Which is the best MQA method?}
Among all the \textit{training-free} methods, we find that \captioning can achieve the best performance and is more robust to different prompts compared to other adaptation methods by an average margin of $3.6\%$ on all styles.
We observe that both utilizing MLLMs for captioning reference images, \summarization suffers more than \captioning adaptation regarding performance and is more sensitive to prompts for all reference styles, which can be observed from the higher standard deviation, showing asking MLLMs to caption and summarize the refinement texts is less controllable. 
To conclude, captioning images is still a golden method since MLLMs and LLMs are powerful enough to generate faithful captions. 

\textbf{Notably}, \sumsuit shows a \textit{non-marginal} improvement ($4.3\%\text{-}9.7\%$) and more \textit{stable} performance across all backbone models. This proves the efficacy and transferability of our SUIT strategy.
To verify our motivation that training-free MQA can output uncontrollable text queries that have a distribution shift from the original NLQs on which the backbones are trained, we visualize the embeddings of original NLQs and adapted MQs in Fig.~\ref{fig:vis} with t-SNE~\cite{van2008visualizing}. It shows that original NLQs have similar distributions as \sumsuit other than the other 2 training-free methods for all different image styles. 

However, the performance gap between our MQ setting and the original NLQ benchmark (refer to Appx.~\ref{sec:origin}) is still remarkable, which shows that the query semantics are more or less distorted across modalities. 

\vspace{0.2cm} 
\noindent 
\textbf{Across different backbone models,} we find that
models that perform well in one adaptation method tend to perform well in others. For example, UVCOM and TR-DETR consistently show high performance across \captioning, \summarization, and \visquery methods. We observe that more recent models keep their outperforming performance on our \benchmark.
Latest models, including UVCOM, TR-DETR, and CG-DETR, tend to perform better across different adaptation methods and reference image styles. In contrast, older models like Moment-DETR consistently show lower performance.
LLM-based models cannot compete with other specialized models without exception; this aligns with their subpar performance on NLQ-based benchmarks~\cite{yu2024self,huang2024vtimellm,ren2023timechat}.  
In the next section, we find that model performance on \benchmark highly correlates with that on natural language query-based benchmark QVHighlights. This shows that (1) our multimodal queries share semantics with the original benchmark; (2) the adaptation methods and models could understand semantics from multimodal queries.

\subsection{Ablation Studies}
Besides the benchmark, we conduct additional studies for other intriguing questions in this section and in Appx.~\ref{sec:moreresults}.

\vspace{0.2cm} 
\noindent 
\textbf{Do Artifacts in synthetic reference images distort the conclusion?}\label{sec:control}
The artifacts in our generated data are inevitable even with the best commercial Text-to-Image models so far. To understand the impact of generated images' artifacts on model evaluation, we conduct a controlled experiment by collecting a subset of MQs by crawling similar images via the Google image search engine. Each image in this retrieved subset has a corresponding generated reference image in a subset $D_{gen}$ of ICQ-Highlight. The retrieval criterion is that retrieved images should be as similar as possible to the generated images in semantics/style/details so that the generation artifacts are the only control variable. The final subset comprises 84 samples from 4 styles.  We compare the model performance on $D_{ret}$ and $D_{gen}$. Our pre-assumption is that if generation artifacts degrade the model performance largely, then $D_{ret}$ should perform better than $D_{gen}$. Otherwise, $D_{gen}$ should perform close to $D_{ret}$.
As shown in Fig.~\ref{fig:control}, model performance on $D_{gen}$ is close to $D_{ret}$ in general. This shows that generation artifacts do not skew our findings largely, and our benchmark is still generalizable.


\vspace{0.2cm} \noindent \textbf{Importance of Refinement Texts}
To assess the impact of refinement texts on video event localization using MQs, we have evaluated model performance using only reference images as queries, omitting refinement texts.
We employ the \captioning adaptation without a modifier for integrating refinement texts. 
As shown in Tab.~\ref{tab:wo_refinement}, we present the model performance and their relative performance drop in percentage compared to those with refinement texts. 
Models have different scales of performance drop, which indicates that refinement texts help refine the semantics of reference images and localize the events.
Additionally, we observe that for \scribble images, the performance drop is less pronounced compared to other styles in that these images are inherently minimalistic and less reliant on details.

%% file: sec/5_conclusion.tex
\section{Conclusion}

    

\vspace{0.2cm}
\noindent
\textbf{Societal Impacts} Using multimodal semantic queries for video event localization brings prospects in real-world applications, such as assisting illiterate, pre-literate, or non-speakers in cross-lingual situations, as it allows them to interact with videos through images as a more accessible and convenient approach.

In this work, we introduce a new benchmark, \benchmark, marking an initial step towards using multimodal semantic queries for video event localization. We have found that our proposed MQA and SUIT methods can accommodate conventional models to MQs effectively, serving as effective baselines for this novel setting. 
Our findings confirm that using MQs for video event localization is practical and feasible.
Nonetheless, the field remains open to innovative model architectures and training paradigms for MQs. We believe our work paves the way for real-world applications that leverage MQs to interact with video content.

%% file: sec/X_suppl.tex
\clearpage
\setcounter{page}{1}

\appendix
\section*{Appendix}

In this Appendix, we present the following:

\begin{itemize}
    \item Additional information about the dataset \dataset and licenses for the datasets and models we have used;
    \item Additional technical implementations including prompts of the benchmark \benchmark;
    \item Extended experimental results due to page limits in the main part.
\end{itemize}

\begin{figure*}[htbp]
    \centering
    \includegraphics[width=0.72\textwidth]{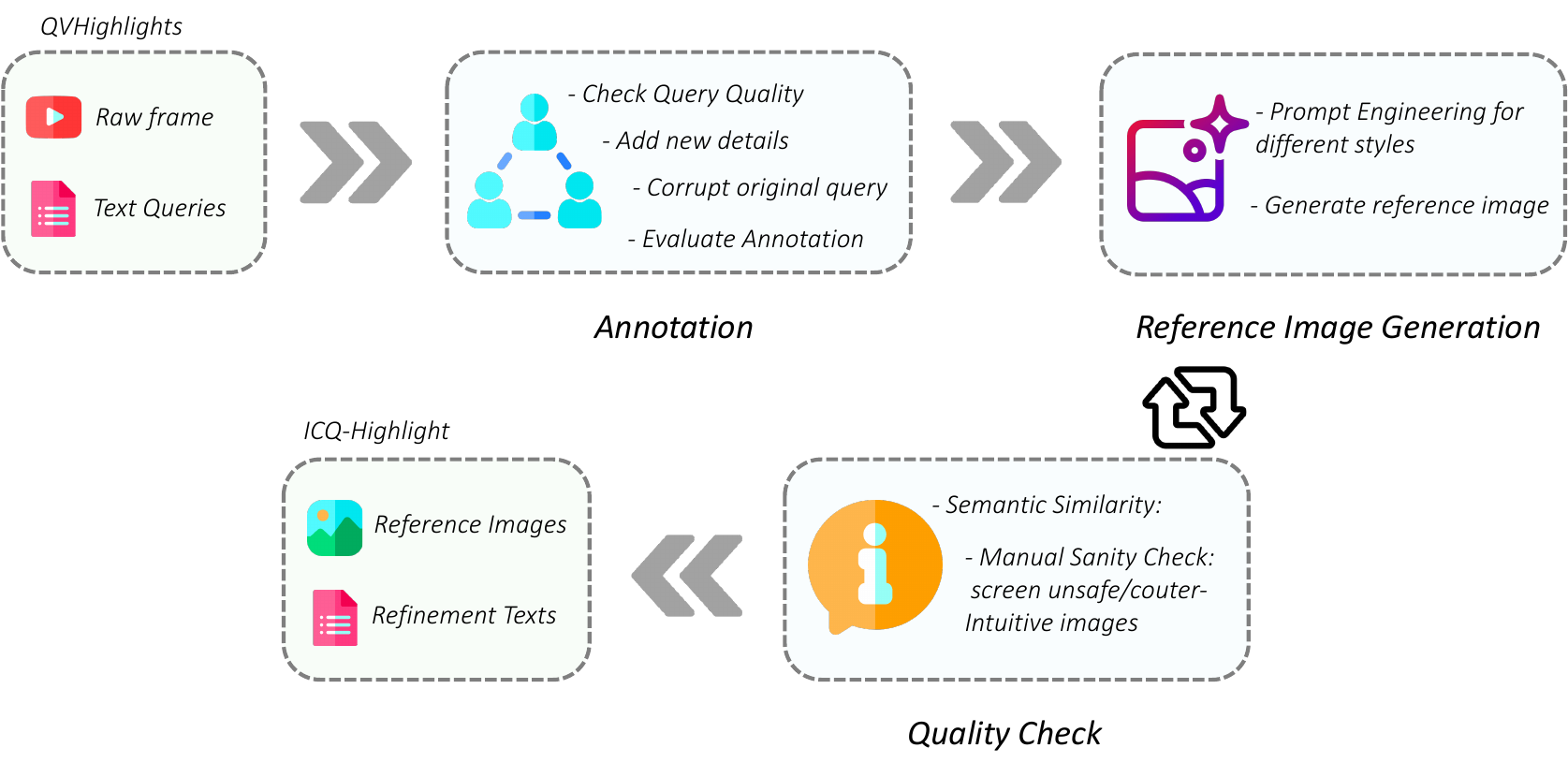}
    \caption{\textbf{Dataset Construction Pipeline:} We base our model with original annotations from QVHighlights and introduce a pipeline consisting of annotation, reference image generation, and quality check.}
    \label{fig:dataset_pipeline}
\end{figure*}

\section{Dataset: ICQ-Highlight}\label{sec:data-app}
\subsection{License}
The dataset and code are publicly accessible. We use standard licenses from the community and provide the following links to the non-commercial licenses for the datasets we used in this paper. 

\textbf{QVHighlights}: \url{https://github.com/jayleicn/moment_detr/blob/main/data/LICENSE}

\textbf{Stability Diffusion}:\url{https://github.com/Stability-AI/stablediffusion/blob/main/LICENSE}

\subsection{Construction Pipeline}
We base our model on the original annotation from QVHighlights~\cite{lei2021detecting}. The whole pipeline, as shown in Fig.~\ref{fig:dataset_pipeline} consists of (1) \textit{annotation}: We further conduct a quality check on the annotations in the original dataset and filter out a few samples (details can be found in Sec.~\ref{sec:delete}). In order to generate more relevant reference images, we manually augment the original captions by adding new visual details based on three frames extracted from the raw videos. To introduce refinement texts, we purposely alter certain details of the captions to generate a new one. All annotations are carried out by two individuals and evaluated by a third party for accuracy. (2) We use the augmented and altered captions to generate reference images with a suite of Text-2-Image models, including DALL-E 2 and Stability Diffusion XL for 4 variants of styles. (3) We implement an additional quality check process for all generated images to eliminate and regenerate images that might contain unsafe or counterintuitive content. We employ BLIP2~\cite{li2023blip} to filter out generated images with lower semantic similarity with augmented captions than 0.2 and conduct a manual sanity check to control the image quality.

\vspace{0.2cm} 
\noindent 
\textbf{Data Curation and Quality check}
Image generation can suffer from significant imperfections in terms of semantic consistency and content safety. To address these issues, we implement a quality check in 2 stages: (1) We calculate the semantic similarity between the generated images and the text queries using BLIP2~\cite{li2023blip} encoders, eliminating samples that score lower than 0.2; (2) We perform human sanity check to replace images that are: i) semantically misaligned with the text, ii) mismatched with the required reference image style, iii) containing sensitive or unpleasant content (\eg, violent, racial, sexual content), counterintuitive elements, or noticeable generation artifacts.

\subsection{Statistics}\label{sec:stats}
The dataset comprises 1515 videos and 1546 test samples on average for each style. The exact numbers may vary slightly across styles and are provided in the Appendix.

Tab.~\ref{tab:stat_dataset} presents the statistics for various reference image styles in terms of the number of queries, videos, and the presence of refinement texts. Tab.~\ref{tab:stat_type_visual} breaks down the statistics of refinement texts for different reference image styles across various query types: object, action, relation, attribute, environment, and others. 
The numbers of each type can vary slightly depending on the different styles.

\begin{figure*}
    \centering    \includegraphics[width=0.72\textwidth]{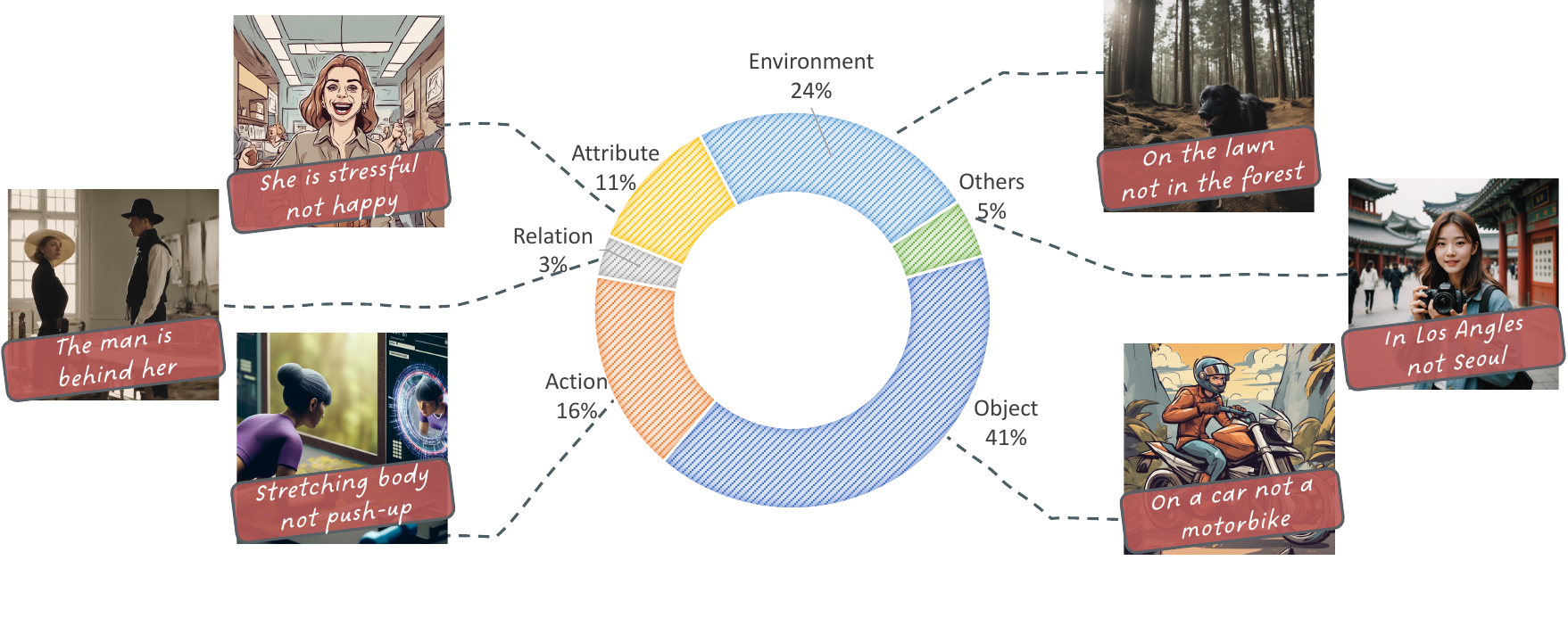}
    \caption{\textbf{Distribution of Refinement Text Types.} Refinement texts are designed to either \textit{complement} or \textit{correct} the original semantics of reference images. We identify 5 major types of refinement texts, each targeting different semantic aspects: object, action, relationship, attribute, environment, and others.}
    \label{fig:dist}
\end{figure*}

\begin{table}[htbp]
\scriptsize
\setlength{\tabcolsep}{2pt}
\centering
\begin{tabular}{ccccc}
\toprule
\begin{tabular}{c} \textbf{Reference} \\ \textbf{Image Style} \end{tabular} & \textbf{\#Queries} &\textbf{\#Videos} & \begin{tabular}{c} \textbf{\#With} \\ \textbf{Refinement Texts} \end{tabular} & \begin{tabular}{c} \textbf{\#Without} \\ \textbf{Refinement Texts} \end{tabular}\\
\midrule
\scribble &  1546 & 1515 & /  & 5\\
\cinematic & 1532 & 1502 & 1445 & 5\\
\cartoon &  1532 & 1501 & 1444 & 5\\
\realistic & 1532 & 1501 & 1446  & 4\\
\bottomrule
\end{tabular}
\caption{Statistics of Different Reference Image Styles}
\label{tab:stat_dataset}
\end{table}

\begin{table}[htbp]
\centering
\scriptsize
\setlength{\tabcolsep}{3pt}
\begin{tabular}{ccccccc}
\toprule
\multirow{2}{*}{\begin{tabular}{c} \textbf{Reference} \\ \textbf{Image Style} \end{tabular}} & \multicolumn{6}{c}{\textbf{\#Queries}}\\
&  \text{Object} & \text{Action} & \text{Relation} & \text{Attribute} & \text{Environment} & \text{Others} \\
\midrule
\scribble & 594 & 242 & 50 & 162 & 343 & 70 \\
\cinematic & 588 & 239 & 50 & 162 & 343 & 66 \\
\cartoon & 590 & 239 & 48 & 161 & 341 & 68 \\
\realistic & 586 & 241 & 50 & 161 & 341 & 70 \\
\bottomrule
\end{tabular}
\caption{Statistics of Refinement Texts}
\label{tab:stat_type_visual}
\end{table}

\begin{table*}
    \centering
    \scriptsize
\setlength{\tabcolsep}{8pt}
    \begin{tabular}{cccccc}
    \toprule
        \textbf{Model} &\textbf{ Visual Encoder }&  \textbf{Query Encoder} &
        \textbf{Localization Decoder$^{*}$}  &
        \textbf{Source} \\
    \midrule
        Moment-DETR (2021) & ViT-B/32 + SlowFast & CLIP Text & DETR & V\\
        QD-DETR (2023) & ViT-B/32 + SlowFast & CLIP Text & DETR & V, V+A\\
        EaTR (2023)& ViT-B/32 + SlowFast & CLIP Text & DETR &  V\\
        CG-DETR (2023)& ViT-B/32 + SlowFast & CLIP Text & DETR & V\\
        TR-DETR (2024)& ViT-B/32 + SlowFast & CLIP Text & DETR & V, V+A\\
        UMT (2022) & ViT-B/32 + SlowFast & CLIP Text & Transformer  & V+A\\
        UniVTG  (2023)& ViT-B/32 + SlowFast & CLIP Text & Conv. Heads  & V \\
        UVCOM  (2023)& ViT-B/32 + SlowFast & CLIP Text & Transformer Heads & V, V+A\\
        SeViLA  (2023)& ViT-G/14 & CLIP Text & Glan-T5 XL (3B)~\cite{chung2024scaling}& V\\
        TimeChat (2024) & ViT-G/14 + Video Q-Former & LLaMA tokenizer & LLaMA-2 7B~\cite{genai2023llama} & V\\
        VTimeLLM (2024) & ViT-L/14 & LLaMA tokenizer & Vicuna v1.5 (7B)~\cite{chiang2023vicuna} & V\\
    \bottomrule
    \end{tabular}
    \caption{\textbf{Comparison of selected baseline models.} $^{*}$We only list the model head for the localization task if the model has multiple heads for different tasks.}
    \label{tab:model_comp}
\end{table*}

\subsection{Details of Deleted Data}~\label{sec:delete}
We removed four entries from the QVHighlight dataset that could cause violent, sexual, sensitive, or graphic content in generation in the original natural language query as listed:
\begin{itemize}
    \item ``A graph depicts penis size.'' (qid: 9737)
    \item ``People mess with the bull statues testicles.'' (qid: 7787) 
    \item ``People butcher meat from a carcass.'' (qid: 4023) 
    \item ``Woman films herself wearing black lingerie in the bathroom.'' (qid: 7685)
\end{itemize}

\section{Benchmark Details}
In this section, we list the details of our selected backbone models, the implementation of our training-free MQA methods, and \suit\xspace strategy.

\subsection{Implementation Details}\label{sec:append_imple}
\paragraph{Automatic Pseudo-MQs Construction} We build the pseudo-MQ dataset from image-text datasets Flickr30K and COCO. We generate captions for the COCO dataset with BLIP-2~\cite{li2023blip}. To forge the original captions, we employ GPT3.5 to process the pure-text captions of each image with the prompts shown in Tab.~\ref{tab:prpt}. 
For each sample, we randomly select one template and refinement text type to generate a forged caption and the corresponding forged part as a refinement text.
In total, we construct a pseudo-MQ dataset with $\num{89420}$ samples for training and $\num{4785}$ samples for validation.

\paragraph{Implementation of \suit} We apply LoRA to all linear layers in the language model of LLaVA-mistral-1.6 with $\text{rank}=32$ and $\text{alpha}=64$ with one epoch on the full dataset. The training takes up to 16 hours on a single NVIDIA A40 GPU. 

\subsection{Model Comparison}\label{sec:compare}
Tab.~\ref{tab:model_comp} compares our selected baseline models. The query encoder denotes the text encoder of each model used to encode natural language queries. Source represents the modalities of the source data, while V and A refer to ``Video'' and ``Audio'' respectively. All models have been fine-tuned on QVHilights.

\subsection{Prompt Engineering}

Since the performance may highly depend on the wording in a prompt, we use 3 different prompts for \captioning and \summarization adaptation methods. In Tab.~\ref{prompts}, the prompts are divided into ``Prompts For Style \cartoon/\cinematic/\realistic'' and ``Prompts for \scribble''. 
This distinction arises because refining \scribble images with complementary texts involves adding new details, slightly differing from other scenarios. Despite this minor variation, the prompt style remains consistent, simulating 3 different user query styles.

For \sumsuit, we use the \textbf{same} prompts as \summarization in the parameter-efficient fine-tuning with LoRA.

\begin{table*}[htbp]
    \centering
    \scriptsize
    \setlength{\tabcolsep}{5pt}
    \begin{tabular}{m{0.1cm}m{6cm}m{6cm}}
        \toprule
        & \textbf{Prompts For Style \cartoon/\cinematic/\realistic} & \textbf{Prompts For Style \scribble} \\
        \midrule
        1 & 
        I have a caption \{INPUT DATA\}, adjust the \{MODIFICATION TYPE\} from \{MODIFIED DETAIL\} to \{ORIGINAL DETAIL\}. The revised caption should remain coherent and logical without introducing any additional details. & 
        I have a caption \{INPUT DATA\}. Modify it by adding \{NEW TYPE\} \{NEW DETAIL\}. The revised caption should remain coherent and logical without introducing additional details. \\
        \hline
        2 & 
        Read this \{INPUT DATA\}! Change the \{MODIFICATION TYPE\} from \{MODIFIED DETAIL\} to \{ORIGINAL DETAIL\}. Then, write a new caption that fits and doesn’t add new stuff. Only give the caption, no extra words. & 
        Read this \{INPUT DATA\}! Add the \{NEW TYPE\} \{NEW DETAIL\} to it. Then, write a new caption that fits and doesn’t add new stuff. Only give the caption, no extra words. \\
        \hline
        3 & 
        Here’s a caption \{INPUT DATA\}. Can you change \{MODIFICATION TYPE\} from \{MODIFIED DETAIL\} to \{ORIGINAL DETAIL\}? After that, make a new caption that makes sense and doesn’t add anything extra. Just write the caption; no explanations are needed. & 
        Here’s a caption \{INPUT DATA\}. Can you add \{NEW TYPE\} \{NEW DETAIL\}? After that, make a new caption that makes sense and doesn’t add anything extra. Just write the caption, no explanations needed. \\
        \bottomrule
    \end{tabular}
    \caption{\textbf{Prompts for \captioning and \summarization.} We use 3 different prompts and report the average performance and standard derivation in other tables.}
    \label{prompts}
\end{table*}

\section{Extended Results}
Due to the page limits, we appended additional experiments and analyses in this section.

\subsection{Main Results for Other Metrics}\label{sec:moreresults}
\begin{figure}
    \begin{subfigure}{\linewidth}
        \centering
    \includegraphics[width=0.94\linewidth]{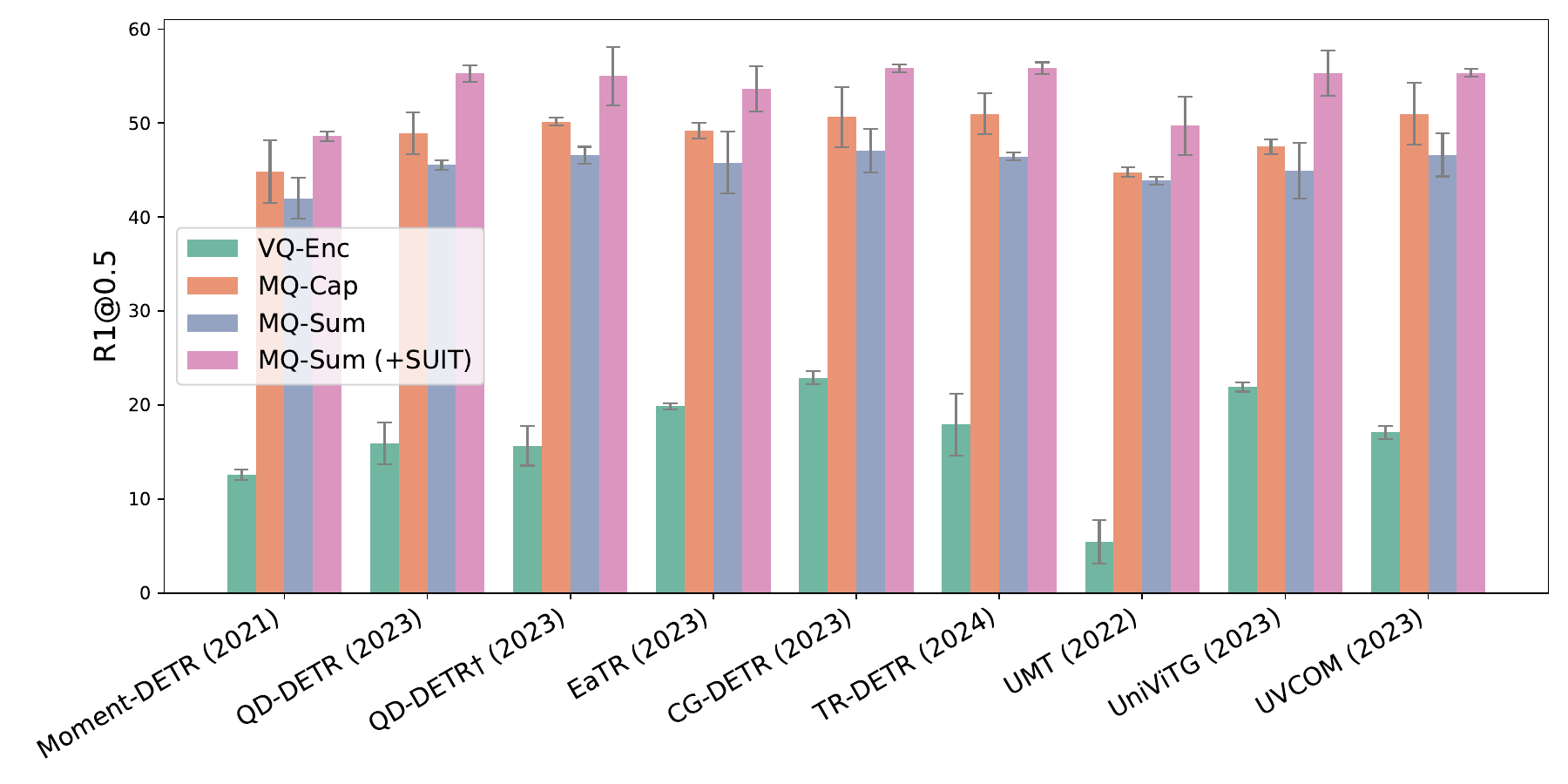}
        \caption{}
    \end{subfigure}
    \vfill
    \begin{subfigure}{\linewidth}
        \centering
        \includegraphics[width=0.94\linewidth]{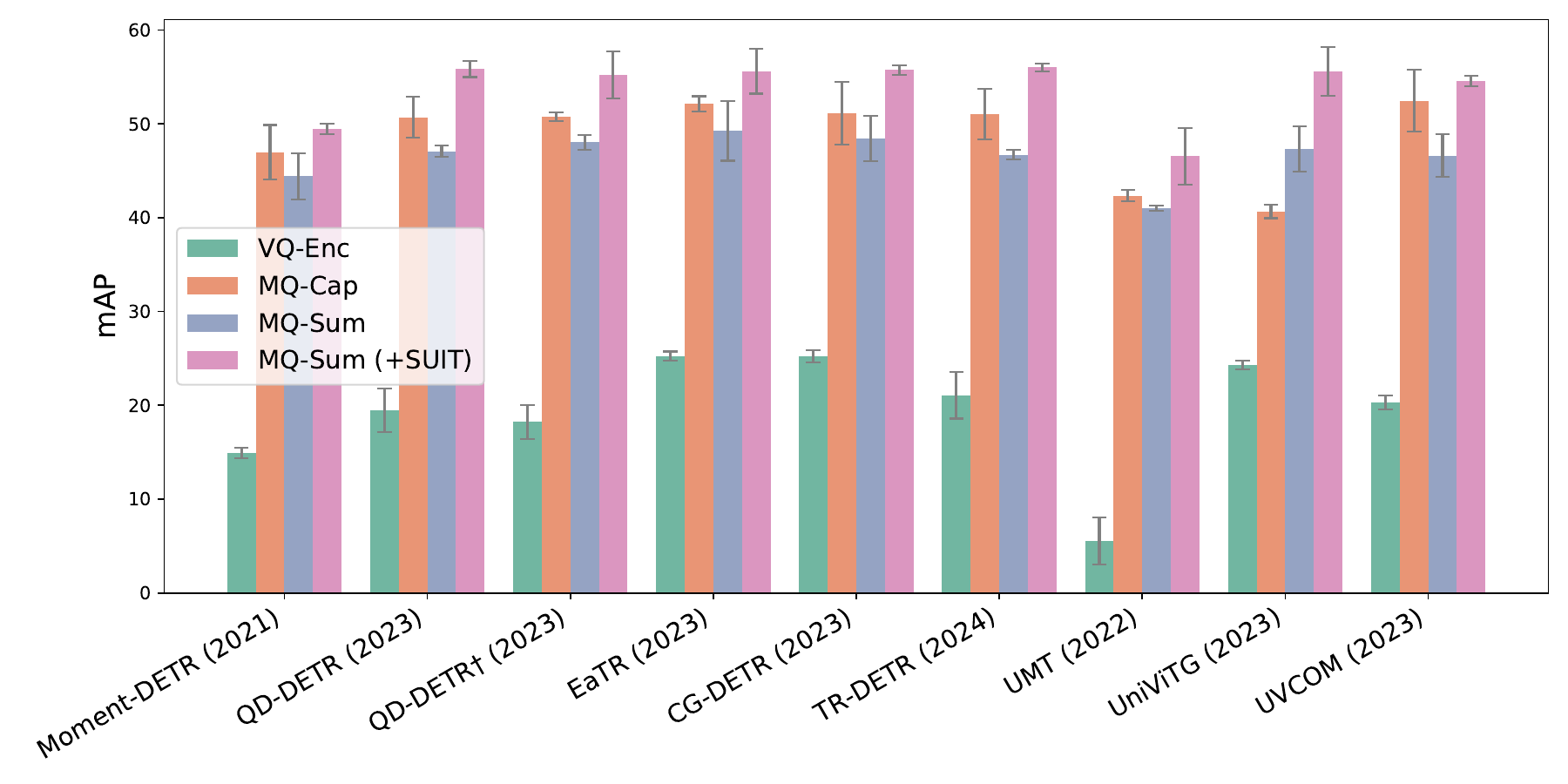}
        \caption{}
    \end{subfigure}
    \caption{Model performance between different MQA methods on \scribble.}
    \label{fig:mqa-compare}
\end{figure}

We present the model performance in mAP in Tab.~\ref{tab:main-results-map} as an extension to Table ~\ref{tab:main-results-recall}. We find that the table aligns with the results stated in Sec.~\ref{sec:ana}. Our \suit\xspace strategy demonstrates good transferability to \dataset.
We highlight this in Fig.~\ref{fig:mqa-compare} on \scribble images and show the performance gain with \sumsuit method.

\begin{table*}[t]
\scriptsize
\setlength{\tabcolsep}{8pt}
\centering
\begin{adjustbox}{center}
\begin{NiceTabular}{llcccccccc}[colortbl-like]
\CodeBefore
\rowcolor{gray!15}{3-8,12-17,21-23,30-32,37-42}
\rowcolor{white}{9-11}
\columncolor{white}{1}
\Body
\toprule
\multirow{2}{*}{\textbf{}} & \multirow{2}{*}{\textbf{Model}}
& \multicolumn{2}{c}{\textbf{\texttt{scribble}}} & 
\multicolumn{2}{c}{\textbf{\texttt{cartoon}}} &
\multicolumn{2}{c}{\textbf{\texttt{cinematic}}}
& \multicolumn{2}{c}{\textbf{\texttt{realistic}}} \\
& & mAP@0.5 & Avg. & mAP@0.5 & Avg. & mAP@0.5 & Avg. & mAP@0.5 & Avg. \\
\midrule
\multirow{9}{*}{\rotatebox{90}{\visquery}}
& Moment-DETR (2021) & 14.95 & 6.67 & 16.51 & 7.21 & 17.00 & 7.39 & 17.41 & 7.66 \\
& QD-DETR (2023) & 19.48 & 10.11 & 19.57 & 10.18 & 18.07 & 9.54 & 18.88 & 9.94 \\
& QD-DETR$\dag$ (2023) & 18.22 & 9.74 & 14.31 & 7.30 & 15.18 & 7.45 & 14.71 & 7.66 \\
& EaTR (2023) & 25.27 & 13.98 & 25.95 & 14.21 & 26.83 & 14.70 & 26.65 & 14.49 \\
& CG-DETR (2023) & \textit{30.24} & \textit{15.57} & \textit{30.78} & \textit{15.70} & \textit{30.07} & \textit{15.48} & \textit{30.98} & \textit{15.83} \\
& TR-DETR (2024) & 21.09 & 11.67 & 20.87 & 11.71 & 19.62 & 11.02 & 19.72 & 10.76 \\
& UMT$\dag$ (2022) & 5.57 & 2.81 & 4.66 & 1.96 & 5.60 & 2.46 & 4.59 & 2.23 \\
& UniVTG (2023) & 24.30 & 13.02 & 20.80 & 11.56 & 19.85 & 10.99 & 19.42 & 10.95 \\
& UVCOM (2023) & 20.13 & 11.15 & 20.19 & 11.96 & 20.67 & 12.37 & 20.73 & 12.03 \\
\midrule
\multirow{10}{*}{\rotatebox{90}{\captioning}}
& Moment-DETR (2021) & 46.98 \rel{± 2.3} & 26.15 \rel{± 1.5} & 48.14 \rel{± 1.2} & 27.22 \rel{± 0.7} & 48.98 \rel{± 0.4} & 27.96 \rel{± 0.4} & 49.00 \rel{± 0.82} & 27.72 \rel{± 0.5} \\
& QD-DETR (2023) & 50.69 \rel{± 3.1} & 31.01 \rel{± 2.4} & 54.15 \rel{± 0.9} & 33.04 \rel{± 0.9} & 55.32 \rel{± 0.9} & 34.06 \rel{± 0.7} & 54.75 \rel{± 0.7} & 34.31 \rel{± 0.7} \\
& QD-DETR$\dag$ (2023) & 50.78 \rel{± 3.9} & 31.44 \rel{± 3.0} & 53.91 \rel{± 1.2} & 33.94 \rel{± 1.0} & 54.06 \rel{± 0.5} & 34.67 \rel{± 0.3} & 53.82 \rel{± 0.8} & 34.18 \rel{± 0.7} \\
& EaTR (2023) & \textit{52.11 \rel{± 2.8}} & 32.88 \rel{± 2.6} & 53.23 \rel{± 0.7} & 33.60 \rel{± 0.7} & 54.00 \rel{± 0.7} & 34.54 \rel{± 0.3} & 54.36 \rel{± 0.8} & 34.73 \rel{± 0.3} \\
& CG-DETR (2023) & 51.13 \rel{± 3.0} & 32.13 \rel{± 2.1} & \textit{56.15 \rel{± 0.8}} & 36.08 \rel{± 0.6} & 55.15 \rel{± 1.0} & 35.22 \rel{± 0.7} & \textit{56.63 \rel{± 0.8}} & 36.57 \rel{± 0.9} \\
& TR-DETR (2024) & 51.07 \rel{± 2.5} & 32.15 \rel{± 2.1} & 55.72 \rel{± 1.1} & 35.98 \rel{± 1.2} & 55.87 \rel{± 0.8} & 36.29 \rel{± 0.5} & 56.32 \rel{± 0.4} & 36.76 \rel{± 0.5} \\
& UMT$\dag$ (2022) & 42.35 \rel{± 2.7} & 26.47 \rel{± 2.0} & 45.03 \rel{± 1.3} & 28.64 \rel{± 1.0} & 46.43 \rel{± 0.8} & 30.01 \rel{± 0.7} & 45.93 \rel{± 0.8} & 29.67 \rel{± 0.8} \\
& UniVTG (2023) & 40.68 \rel{± 2.5} & 24.71 \rel{± 1.9} & 42.68 \rel{± 0.7} & 26.03 \rel{± 0.6} & 43.53 \rel{± 0.4} & 26.43 \rel{± 0.5} & 43.64 \rel{± 0.8} & 26.76 \rel{± 0.5} \\
& UVCOM (2023) & 51.27 \rel{± 3.2} & \textit{33.39 \rel{± 2.5}} & 54.40 \rel{± 0.7} & \textit{36.50 \rel{± 0.7}} & \textit{55.99 \rel{± 0.7}} & \textit{37.11 \rel{± 0.3}} & 54.98 \rel{± 0.8} & \textit{36.83 \rel{± 0.6}} \\
& SeViLA (2023) & 14.45 \rel{± 0.8}& 9.30 \rel{± 0.6}& 19.52 \rel{± 0.5}& 13.12 \rel{± 0.4}& 22.16 \rel{± 0.3}& 14.64 \rel{± 0.4} & 22.48 \rel{± 0.6}& 14.55 \rel{± 0.5}\\
& TimeChat (2024) & 9.08 \rel{± 0.6} & 4.45 \rel{± 0.4} & 11.01 \rel{± 0.9} & 5.13 \rel{± 0.5} & 10.58 \rel{± 0.7} & 4.82 \rel{± 1.0} & 10.69 \rel{± 1.0} & 4.78 \rel{± 0.2}\\
& VTimeLLM (2024) & 18.48 \rel{± 1.0} & 8.15 \rel{± 0.5} & 21.90 \rel{± 0.3} & 9.16 \rel{± 0.1} & 24.03 \rel{± 
 0.5} & 10.15 \rel{± 0.3} & 23.45 \rel{± 0.7} & 10.10 \rel{± 0.1} \\
\midrule
\multirow{10}{*}{\rotatebox{90}{\summarization}}
& Moment-DETR (2021) & 44.40 \rel{± 2.5} & 23.96 \rel{± 1.8} & 47.31 \rel{± 2.1} & 26.03 \rel{± 1.4} & 46.62 \rel{± 1.9} & 25.55 \rel{± 1.3} & 47.29 \rel{± 2.2} & 26.07 \rel{± 1.3} \\
& QD-DETR (2023) & 47.09 \rel{± 2.8} & 28.27 \rel{± 2.4} & 51.06 \rel{± 3.3} & 30.90 \rel{± 2.5} & 50.89 \rel{± 3.3} & 30.52 \rel{± 2.8} & 50.05 \rel{± 3.6} & 30.49 \rel{± 2.7} \\
& QD-DETR$\dag$ (2023) & 48.10 \rel{± 3.2} & 29.49 \rel{± 2.9} & 50.72 \rel{± 3.3} & 31.11 \rel{± 3.0} & 49.94 \rel{± 2.8} & 31.38 \rel{± 2.4} & 50.30 \rel{± 3.8} & 30.85 \rel{± 2.6} \\
& EaTR (2023) & \textit{49.07 \rel{± 2.6}} & \textit{30.92 \rel{± 2.0}} & 50.82 \rel{± 2.6} & 31.38 \rel{± 1.7} & 50.71 \rel{± 3.2} & 31.34 \rel{± 2.7} & 51.37 \rel{± 3.0} & 32.02 \rel{± 2.0} \\
& CG-DETR (2023) & 48.41 \rel{± 3.5} & 29.86 \rel{± 2.9} & 52.31 \rel{± 2.9} & 33.21 \rel{± 2.3} & 51.59 \rel{± 2.8} & 32.34 \rel{± 2.5} & 52.31 \rel{± 3.1} & 32.91 \rel{± 2.0} \\
& TR-DETR (2024) & 46.69 \rel{± 3.6} & 29.72 \rel{± 2.8} & \textit{52.41 \rel{± 2.6}} & 33.48 \rel{± 1.9} & \textit{52.39 \rel{± 3.1}} & 33.14 \rel{± 2.6} & \textit{52.87 \rel{± 3.1}} & \textit{33.57 \rel{± 2.5}}\\
& UMT$\dag$ (2022) & 40.99 \rel{± 2.7} & 25.88 \rel{± 1.8} & 43.03 \rel{± 2.0} & 27.02 \rel{± 1.5} & 42.88 \rel{± 2.0} & 26.73 \rel{± 1.6} & 43.89 \rel{± 1.3} & 27.38 \rel{± 1.0} \\
& UniVTG (2023) & 38.86 \rel{± 2.7} & 22.76 \rel{± 1.8} & 40.13 \rel{± 2.8} & 24.43 \rel{± 1.7} & 40.73 \rel{± 2.7} & 24.02 \rel{± 1.9} & 40.20 \rel{± 2.4} & 24.11 \rel{± 1.6} \\
& UVCOM (2023) & 47.33 \rel{± 3.2} & 30.75 \rel{± 2.5} & 52.22 \rel{± 3.4} & \textit{34.00 \rel{± 2.7}} & 51.37 \rel{± 4.2} & \textit{33.36 \rel{± 3.1}} & 51.64 \rel{± 3.8} & 33.52 \rel{± 2.6} \\
& SeViLA (2023) & 14.54 \rel{± 1.7} & 9.24 \rel{± 1.3}& 22.13 \rel{± 1.8} & 14.07 \rel{± 1.1}& 22.17 \rel{± 1.4}& 14.52 \rel{± 0.9}& 22.87 \rel{± 1.8} & 14.45 \rel{± 1.3}\\
& TimeChat (2024) & 9.12 \rel{± 0.4} & 4.07 \rel{± 0.2} & 9.63 \rel{± 1.7} & 4.64 \rel{± 0.7} & 10.18 \rel{± 1.2} & 4.94 \rel{± 0.9} & 9.46 \rel{± 1.8} & 4.16 \rel{± 1.3}\\
& VTimeLLM (2024) & 19.40 \rel{± 1.4} & 8.54 \rel{± 0.4} & 21.59 \rel{± 0.8} & 8.98 \rel{± 0.4} & 22.74 \rel{± 0.3} & 9.44 \rel{± 0.3} & 23.2 \rel{± 1.6} & 9.65 \rel{± 0.7}\\
\midrule
\multirow{11}{*}{\rotatebox{90}{\summarization}}
& \underline{\textit{\textbf{+ SUIT}}} \\
& Moment-DETR (2021) & 49.46 \rel{± 0.6} & 28.36 \rel{± 0.47} & 49.01 \rel{± 0.3} & 28.0 \rel{± 0.2} & 49.32 \rel{± 0.5} & 28.07 \rel{± 0.3} & 48.39 \rel{± 0.4} & 27.34 \rel{± 0.2} \\
& QD-DETR (2023) & 55.82 \rel{± 0.2} & 35.19 \rel{± 0.1} & 54.12 \rel{± 0.5} & 33.94 \rel{± 0.2} & 55.05 \rel{± 0.2} & 34.59 \rel{± 0.2} & 54.62 \rel{± 0.2} & 34.45 \rel{± 0.2} \\
& QD-DETR$\dag$ (2023) & 54.71 \rel{± 0.5} & 35.29 \rel{± 0.3} & 54.20 \rel{± 0.1} & 35.48 \rel{± 0.2} & 54.05 \rel{± 0.17} & 35.2 \rel{± 0.4} & 53.14 \rel{± 0.6} & 34.54 \rel{± 0.2} \\
& EaTR (2023) & 55.2 \rel{± 0.7} & 35.86 \rel{± 0.4} & 52.88 \rel{± 0.2} & 34.18 \rel{± 0.2} & 54.07 \rel{± 0.7} & 34.66 \rel{± 0.1} & 52.68 \rel{± 0.3} & 33.92 \rel{± 0.4} \\
& CG-DETR (2023) & 55.6 \rel{± 0.6} & 36.16 \rel{± 0.2} & 55.5 \rel{± 0.4} & 35.47 \rel{± 0.3} & 55.93 \rel{± 0.7} & 35.85 \rel{± 0.3} & 55.34 \rel{± 0.6} & 35.43 \rel{± 0.3} \\
& TR-DETR (2024) & \textit{\textbf{56.75}} \rel{± 0.4} & \textit{\textbf{37.25}} \rel{± 0.2} & \textit{\textbf{55.76}} \rel{± 0.2} & \textit{\textbf{36.31}} \rel{± 0.1} & \textit{\textbf{56.36}} \rel{± 0.5} & \textit{\textbf{36.84}} \rel{± 0.5} & \textit{\textbf{56.18}} \rel{± 0.3} & \textit{\textbf{37.05}} \rel{± 0.3} \\
& UMT$\dag$ (2022) & 46.55 \rel{± 0.3} & 30.45 \rel{± 0.3} & 46.44 \rel{± 0.6} & 30.71 \rel{± 0.3} & 46.86 \rel{± 0.4} & 30.9 \rel{± 0.3} & 46.54 \rel{± 0.2} & 29.94 \rel{± 0.2} \\
& UniVTG (2023) & 43.36 \rel{± 0.4} & 26.87 \rel{± 0.2} & 42.2 \rel{± 0.4} & 26.42 \rel{± 0.2} & 43.23 \rel{± 0.5} & 26.81 \rel{± 0.3} & 42.89 \rel{± 0.58} & 26.45 \rel{± 0.4} \\
& UVCOM (2023) & 54.18 \rel{± 0.3} & 36.92 \rel{± 0.4} & 54.56 \rel{± 0.3} & 36.91 \rel{± 0.1} & 54.43 \rel{± 0.4} & 37.29 \rel{± 0.1} & 53.31 \rel{± 0.5} & 36.53 \rel{± 0.2} \\

\bottomrule
\end{NiceTabular}%
\end{adjustbox}
\caption{\textbf{Model performance (mAP) on \benchmark.} 
We highlight the best score in \textit{italic} for each adaptation method and the overall best scores in \textbf{bold}.
For \captioning and \summarization, we report the standard deviation of 3 runs with different prompts and for \sumsuit we report the average performance with different seeds in training. $\dag$ uses extra audio modality.}  
\label{tab:main-results-map}
\end{table*}

\subsection{Model Performance on Different Refinement Text Types}\label{sec:subset}
\begin{figure*}[htbp]
    \centering
    \begin{subfigure}{0.23\textwidth}
        \centering
        \includegraphics[width=1\linewidth]{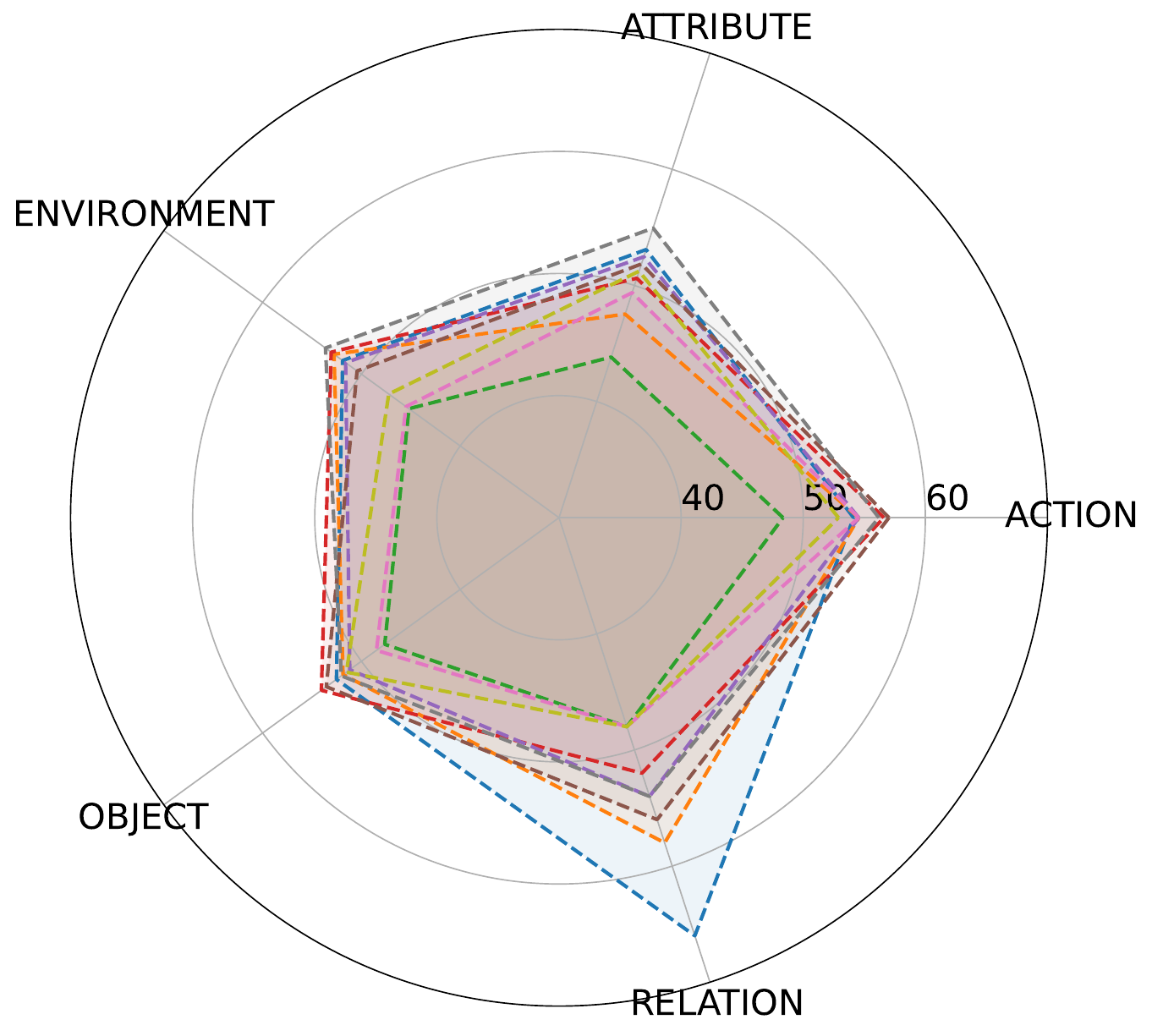}
        \caption{scribble}
    \end{subfigure}
    \hfill
    \begin{subfigure}{0.23\textwidth}
        \centering
        \includegraphics[width=1\linewidth]{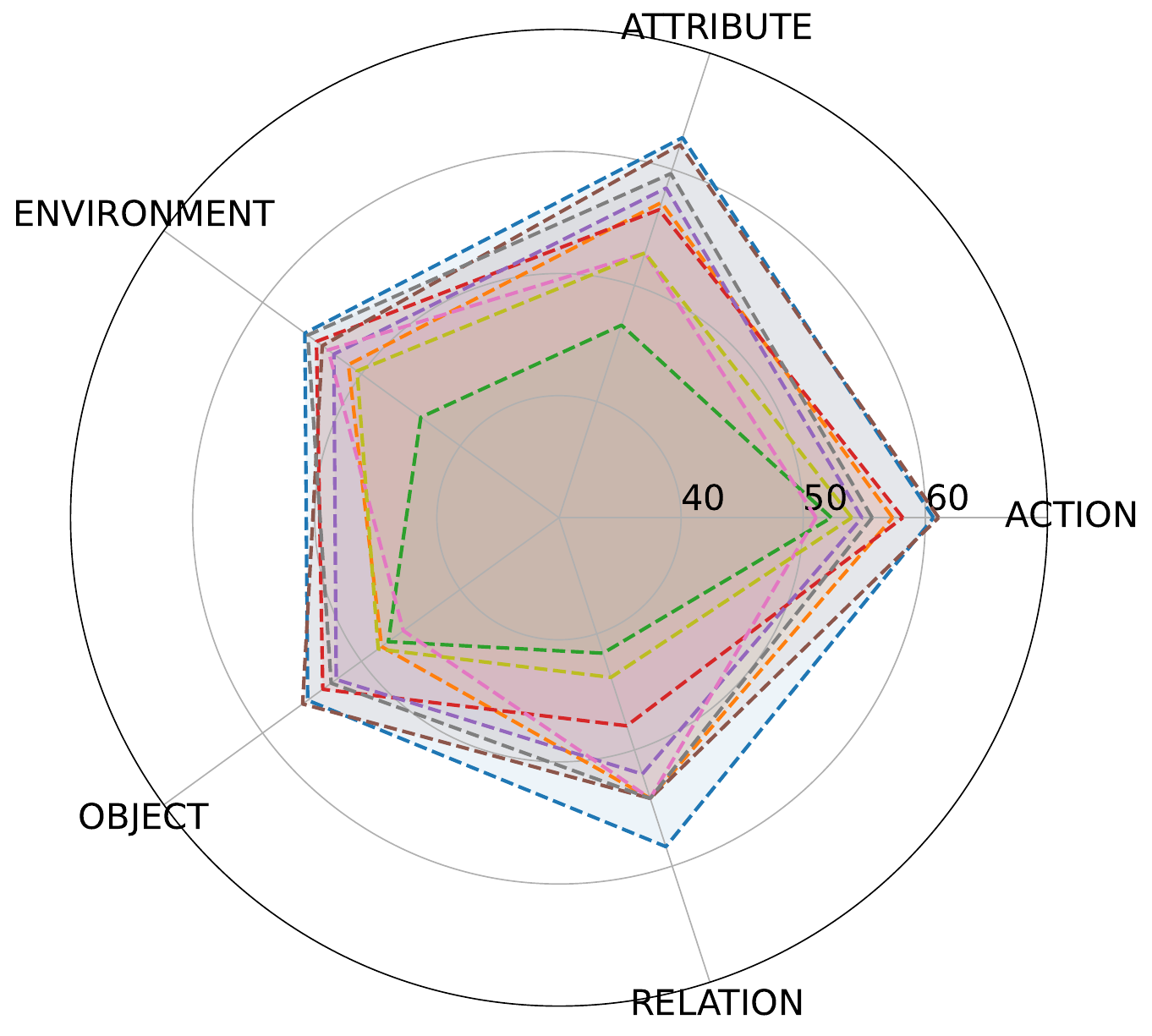}
        \caption{cartoon}
    \end{subfigure}
    \hfill
    \begin{subfigure}{0.23\textwidth}
        \centering
        \includegraphics[width=1\linewidth]{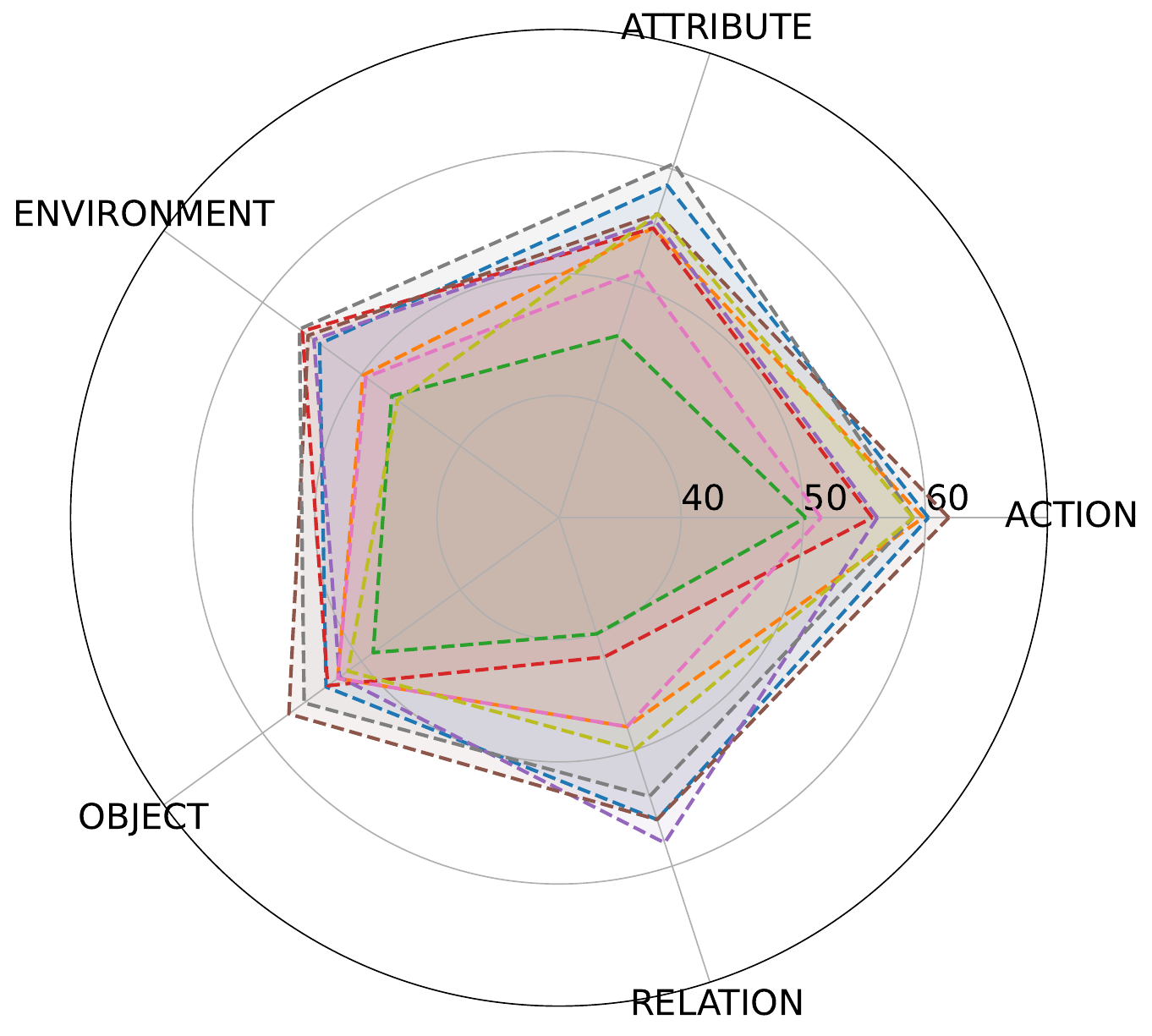}
        \caption{cinematic}
    \end{subfigure}
    \hfill
    \begin{subfigure}{0.23\textwidth}
        \centering
        \includegraphics[width=1\linewidth]{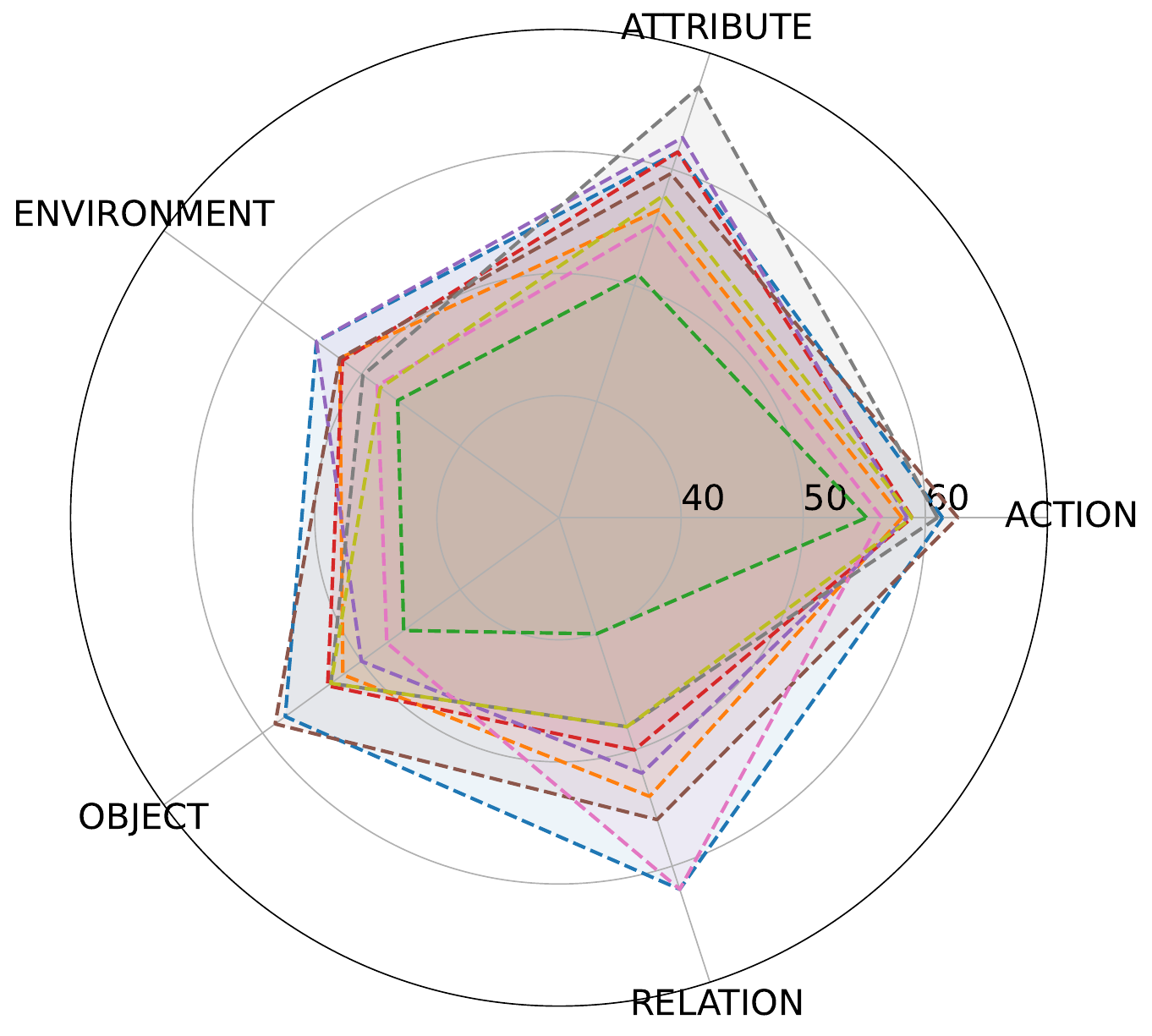}
        \caption{realistic}
    \end{subfigure}
    \caption{\textbf{Model performance on different subsets of refinement text types.} We observe that model performance with different refinement text types varies across styles.}
    \label{fig:subset}
\end{figure*}

We calculate the model performance on different subsets of refinement texts shown in Fig.~\ref{fig:subset}.
We conclude even though models have close performance across reference image styles, they show varied performance on different refinement text types across styles. For \scribble style, models generally perform for ``relation'' better than other styles. For \cartoon style, models demonstrate a more balanced performance across all types. The performance is notably higher for ``environment'' and ``attribute'' in \cinematic style. Finally, for \realistic style, the models yield better performance in ``object'' and ``environment''.

\begin{table*}[htbp]
\scriptsize
\centering
\begin{tabular}{lcccccc}
\toprule
\multirow{2}{*}{\textbf{Method}} & \multicolumn{5}{c}{\textbf{original NLQ (Performance on QVHighlights)}}\\
 & \textbf{R1@0.5} & \textbf{R1@0.7} & \textbf{mAP@0.5} & \textbf{mAP@0.7} & \textbf{Avg.}\\
\midrule 	
Moment-DETR (2021) & 54.92 \tiny{(-4.6$\%$)} & 36.87 \tiny{(-3.3$\%$)}& 55.95 \tiny{(-4.2$\%$)}& 31.59 \tiny{(-4.5$\%$)}& 32.54 \tiny{(-3.8$\%$)}\\
QD-DETR (2023) & 62.87 \tiny{(-8.6$\%$)}& 46.70 \tiny{(-12.5$\%$)} & 62.66 \tiny{(-7.6$\%$)}& 41.59 \tiny{(-12.4$\%$)}& 41.23 \tiny{(-10.3$\%$)}\\
QD-DETR$\dag$ (2023) & 63.71 \tiny{(-6.2$\%$)}& 47.67 \tiny{(-8.1$\%$)}& 62.9 \tiny{(-5.6$\%$)}& 42.07 \tiny{(-6.6$\%$)}& 41.73 \tiny{(-6.4$\%$)}\\
EaTR (2023) & 60.93 \tiny{(-8.0$\%$)}& 46.12 \tiny{(-9.5$\%$)}& 62.01 \tiny{(-5.9$\%$)}& 42.11 \tiny{(-7.6$\%$)}& 41.39 \tiny{(-6.7$\%$)}\\
CG-DETR (2023) & 67.27 \tiny{(-8.9$\%$)}& 51.94 \tiny{(-13.6$\%$)}& 65.48 \tiny{(-7.6$\%$)}& 45.64 \tiny{(-12.4$\%$)}& 44.88 \tiny{(-11.3$\%$)}\\
TR-DETR (2024) & 67.08 \tiny{(-7.5$\%$)}& 51.36 \tiny{(-8.3$\%$)}& 66.20 \tiny{(-7.3$\%$)}& 46.28 \tiny{(-9.3$\%$)}& 44.99 \tiny{(-8.1$\%$)}\\
UMT$\dag$ (2022) & 60.22 \tiny{(-10.0$\%$)}& 44.24 \tiny{(-14.1$\%$)}& 56.62 \tiny{(-9.5$\%$)}& 39.85 \tiny{(-15.2$\%$)}& 38.54 \tiny{(-12.9$\%$)}\\
UniVTG (2023) & 59.70 \tiny{(-8.7$\%$)}& 40.82 \tiny{(-7.2$\%$)}& 51.22 \tiny{(-8.0$\%$)}& 32.84 \tiny{(-9.9$\%$)}& 32.53 \tiny{(-9.0$\%$)}\\
UVCOM (2023) & 65.01 \tiny{(-5.6$\%$)}& 51.75 \tiny{(-8.0$\%$)}& 64.88 \tiny{(-5.3$\%$)}& 46.96 \tiny{(-9.0$\%$)}& 45.83 \tiny{(-8.2$\%$)}\\
SeViLA (2023) & 56.57 \tiny{(-56.2$\%$)}& 40.45 \tiny{(-62.1$\%$)}& 47.14 \tiny{(-56.8$\%$)}& 32.69 \tiny{(-62.3$\%$)}& 33.10 \tiny{(-60.6$\%$)}\\
\bottomrule
\end{tabular}
\caption{Performance comparison between the original NLQ (in QVHighlights) and forged NLQ with refinement texts introduced in \dataset. The performance drop highlighted in the parenthesis indicates that the modifications on natural language query are non-trivial. $\dag$ indicates the usage of additional audio modality.}
\label{tab:ori_vs_mod}
\end{table*}

\subsection{MQ-based \vs NLQ-based Performance}
We compare model performance on the MQ-based \dataset and the original NLQ-based QVHighlight (results taken from the original papers) using Spearman's rank correlation coefficient~\cite{spearman1961proof} on R1@0.5. 
For \scribble, Spearman's rank correlation coefficients are 0.89(\captioning) and 0.93(\summarization). The \cartoon style yields coefficients of 0.98(\captioning) and 0.94(\summarization). The \cinematic style shows coefficients of 0.93 for both \captioning and \summarization. Lastly, \realistic has coefficients of 0.96(\captioning) and 0.95(\summarization). The high correlation scores indicate a strong positive correlation across benchmarks, suggesting queries of both benchmarks share the common semantics and yield the reliability of our benchmark.

\subsection{Captioning Without Refinement Text \vs Visual Query Encoding}

We compare the model performance between \captioning without the revision step with refinement texts and \visquery, as shown in Tab.~\ref{tab:table-cap-wo-text}. Both methods only use reference images as queries without refinement texts. 
Overall, \captioning without refinement texts still significantly outperforms pure \visquery, highlighting the effectiveness of image captioning. Additionally, TR-DETR and UVCOM perform best across all styles.

\subsection{Original NLQs (in QVHighlights) vs. Forged NLQs in \dataset}\label{sec:origin}
We have evaluated the model performance based on the original NLQs in QVHighlights and our refinement texts introduced in MQs to assess the significance of the refinement texts and the sensitivity of different models to natural language queries. ~\cite{moon2023query} points out that the impact of the NLQs may be minimal for some existing models, such as Moment-DETR. As shown in Tab.~\ref{tab:ori_vs_mod}, Moment-DETR exhibits relatively smaller drops across all metrics, supporting this claim. On the other hand, the latest models, such as CG-DETR and TR-DETR, experience more significant performance drops, indicating a higher sensitivity to query modifications. Furthermore, SeViLA is extremely sensitive to query modifications, shown by severe performance declines across all evaluated metrics. Overall, the considerable performance decline across various models demonstrates that our modifications significantly affect the original queries. This also shows that our introduced refinement texts are not semantically trivial for localizing with multimodal queries.

\subsection{Case Study: the Impact of Potential Generation Artifact}
\begin{figure*}
    \centering
    \includegraphics[width=1.\linewidth]{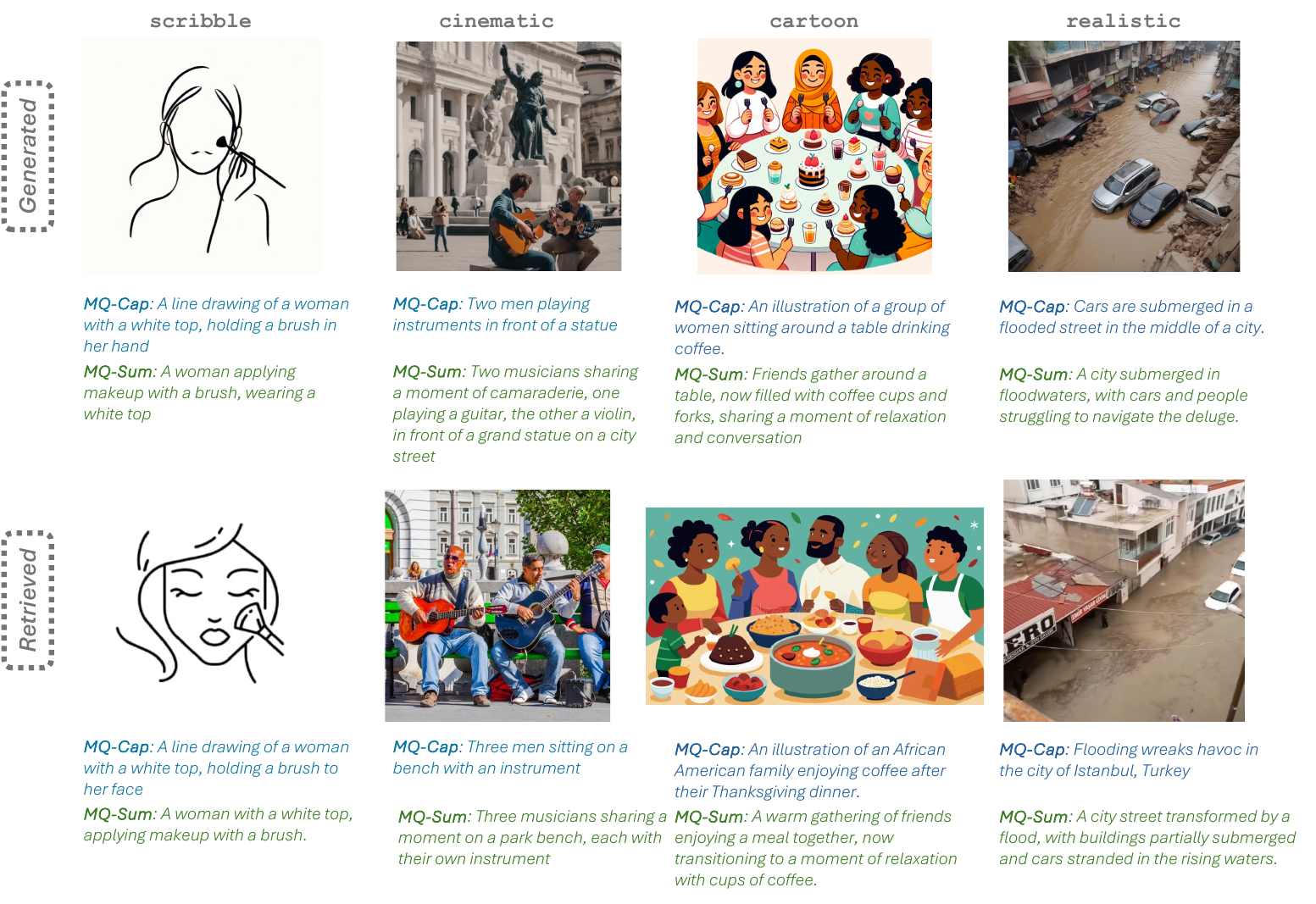}
    \caption{We showcase four examples in our subsets $D_{gen}$ and $D_{ret}$. We notice that image generation artifacts usually do not change the image semantics dramatically and thus do not influence the caption directly. \textit{Please note that the retrieved images provided are for research purposes only. Distribution or sharing of these images without proper authorization is strictly prohibited.}}
    \label{fig:case}
\end{figure*}

Along with the controlled experiment shown in Sec.~\ref{sec:control}, we conduct a qualitative case study with samples in the subsets $D_{gen}$ and $D_{ret}$.
We notice that generation artifacts usually do not change the image semantics and thus do not influence the caption dramatically, as shown in Fig.~\ref{fig:case}.

While collecting this subset, we noticed that AI-generated images become more prevalent on the Internet. This indicates that our generated dataset has a more realistic application and reflects the practical scenarios when users aim to locate events with generated images online. In addition, we find that generation artifacts do not pose significant issues in scribble and cartoon styles since the images are already simple.

\begin{table*}[htbp]
\scriptsize
\setlength{\tabcolsep}{6pt}
\centering
\begin{adjustbox}{center}
\begin{NiceTabular}{llcccccccc}[colortbl-like]
\CodeBefore
\Body
\toprule
\multirow{2}{*}{\textbf{}} & \multirow{2}{*}{\textbf{Model}}
& \multicolumn{2}{c}{\textbf{\texttt{scribble}}} & 
\multicolumn{2}{c}{\textbf{\texttt{cartoon}}} &
\multicolumn{2}{c}{\textbf{\texttt{cinematic}}}
& \multicolumn{2}{c}{\textbf{\texttt{realistic}}} \\
& & R1@0.5 & R1@0.7 & R1@0.5 & R1@0.7 & R1@0.5 & R1@0.7 & R1@0.5 & R1@0.7 \\
\midrule
\multirow{10}{*}{\rotatebox{90}{\captioning wo/ revision}}
& Moment-DETR (2021) &  45.15 & 28.72 & 43.60 & 27.94 & 44.06 & 29.70 &  44.06 & 28.98\\
& QD-DETR (2023) & 49.81 & 33.70 & 49.87 & 34.33 & 49.67 & 34.73 & 50.52 & 35.25\\
& QD-DETR$\dag$ (2023) & 51.29 & 36.03 & 48.69 & 33.88 & 49.48 & 34.99 &  49.93 & 35.05\\
& EaTR (2023) & 52.01 & 37.77 & 47.45 & 33.09 & 48.56 & 34.33 &  49.61 & 35.64\\
& CG-DETR (2023) & 51.42 & 37.84 & 49.35 & 35.90 & 48.89 & 34.79 &  51.04 & 36.55\\
& TR-DETR (2024) & 52.01 & 37.19 & 51.04 & 36.62 & 50.00 & 36.03 &  \textbf{52.28} & \textbf{37.53} \\
& UMT$\dag$ (2022) & 46.25 & 31.57 & 45.82 & 30.61 & 46.34 & 29.96 & 46.08 & 31.85\\
& UniVTG (2023) & 47.87 & 33.76 & 45.56 & 29.24 & 45.43 & 29.05 & 46.80 & 30.42\\
& UVCOM (2023) & \textbf{52.26} & \textbf{39.39} & \textbf{51.50} & \textbf{37.99} & \textbf{50.98} & \textbf{36.75} &  51.70 & \textbf{37.53}\\

\midrule
\multirow{9}{*}{\rotatebox{90}{\textsc{\visquery}}} &
Moment-DETR (2021) & 12.55 & 5.69 & 13.38 & 6.59 & 14.36 & 6.01 & 14.88 & 6.53 \\
& QD-DETR (2023) & 15.91 & 9.12 & 14.88 & 8.62 & 13.90 & 8.49 & 14.62 & 8.36 \\
& QD-DETR$\dag$ (2023) & 15.65 & 10.03 & 12.60 & 6.79 & 12.34 & 6.72 & 12.34 & 7.44 \\
& EaTR (2023) & 19.86 & \textbf{13.00} & 19.91 & 12.99 & 21.15 & \textbf{13.45} & 21.48 & 13.38 \\
& CG-DETR (2023) & \textbf{22.90} & \textbf{13.00} & \textbf{24.93} & 13.58 & \textbf{23.24} & 13.12 & \textbf{24.74} & \textbf{14.23} \\
& TR-DETR (2024) & 17.92 & 11.19 & 17.36 & 11.10 & 15.14 & 9.86 & 15.60 & 9.53 \\
& UMT$\dag$ (2022) & 5.43 & 2.85 & 4.77 & 2.09 & 5.22 & 2.35 & 4.57 & 2.42 \\
& UniVTG (2023) & 21.93 & 13.00 & 23.89 & \textbf{13.64} & 22.78 & 13.19 & 22.52 & 12.79 \\
& UVCOM (2023) & 17.08 & 9.77 & 16.78 & 10.97 & 17.36 & 11.68 & 17.10 & 11.23 \\
\bottomrule
\end{NiceTabular}
\end{adjustbox}
\caption{\textbf{Model performance (Recall) of \captioning without refinement text and \visquery on \benchmark.}
We highlight the best score in \textbf{bold} for both methods and reference image style. }
\label{tab:table-cap-wo-text}
\end{table*}

\begin{table*}
\scriptsize
\setlength{\tabcolsep}{6pt}
\centering
\begin{tabular}{lc}
\toprule
Type & Prompt \\
\midrule
Object & \begin{tabular}{l} In this task, you are given an input sentence. Your job is to generate a sentence with a different meaning by only changing the main entities (subject, \\object, people, animal, ...) in the input sentence, the others remain unchanged, make sure the modified sentence are still reasonable. Only output the \\modified sentence, do not include explanations. Input sentence: ``\{\}". Output: \end{tabular} \\
Attributes & \begin{tabular}{l} In this task, you are given an input sentence. Your job is to generate a sentence with a different meaning by only changing the attributes (such as color, \\ size, shape, texture, ...) of the objects in the input sentence, the others remain unchanged, make sure the modified sentence are still reasonable. Only \\ output the modified sentence, do not include explanations. Input sentence: ``\{\}". Output:  \end{tabular} \\
Actions & \begin{tabular}{l} In this task, you are given an input sentence. Your job is to generate a sentence with a different meaning by only changing the action verbs in the input \\ sentence, the others remain unchanged, make sure the modified sentence are still reasonable. Only output the modified sentence, do not include expla- \\nations. Input sentence: ``\{\}". Output: \end{tabular} \\
Environment & \begin{tabular}{l} In this task, you are given an input sentence. Your job is to generate a sentence with a different meaning by only changing the environment (focus on \\ `where', such as the background, location, atmosphere, settings...) in the input sentence, the others remain unchanged, make sure the modified sentence \\ are still reasonable. Only output the modified sentence, do not include explanations. Input sentence: ``\{\}". Output: \end{tabular} \\
Relations & \begin{tabular}{l} In this task, you are given an input sentence. Your job is to generate a sentence with a different meaning by only changing the relationships (focus on \\ the relationship between different entities, such as spatial, temporal, interaction-based connections, ...) in the input sentence, make sure the modified \\sentence are still reasonable. Only output the modified sentence, do not include explanations. Input sentence:``\{\}". Output: \end{tabular} \\
\bottomrule
\end{tabular}
\caption{Examples of prompt templates used to generate forged captions with GPT3.5.}
\label{tab:prpt}
\end{table*}